\documentclass{article}

\usepackage[nonatbib,preprint]{neurips_2024}

\usepackage[utf8]{inputenc} %
\usepackage[T1]{fontenc}    %
\usepackage[pagebackref=true,breaklinks=true,colorlinks,bookmarks=false]{hyperref}
\usepackage{url}            %
\usepackage{booktabs}       %
\usepackage{amsfonts}       %
\usepackage{nicefrac}       %
\usepackage{microtype}      %
\usepackage[dvipsnames]{xcolor}         %
\usepackage{mathtools}
\usepackage{enumitem}
\usepackage{amsmath,amsfonts,amssymb,amsthm}
\usepackage{pifont}
\usepackage{times}
\usepackage{epsfig}
\usepackage{graphicx}
\usepackage[sort,nocompress]{cite}
\usepackage{multirow}
\usepackage[heightadjust=all]{floatrow}
\usepackage{caption}
\usepackage{subcaption}
\usepackage{array}
\usepackage{wrapfig}
\usepackage{xspace}
\usepackage{amsmath,amsfonts,bm}
\usepackage{ulem}

\usepackage{wrapfig,lipsum,booktabs}

\newcommand{\cmark}{\ding{51}}%

\newcommand{\methodname}{\textbf{SytheOcc}\xspace}
\newcommand{\methodnamesmall}{SytheOcc\xspace}

\newcommand{\myparagraph}[1]{\vspace{3pt}\noindent\textbf{#1}\quad}

\newcommand{\mpiname}[0]{{\fontfamily{txtt}\selectfont {MPI}}\xspace}
\newcommand{\occname}[0]{{\fontfamily{txtt}\selectfont {Occupancy}}\xspace}

\usepackage{slashbox}

\usepackage{floatrow}
\newfloatcommand{capbtabbox}{table}[][\FBwidth]

\usepackage{makecell}
\definecolor{barrier}{RGB}{255, 158, 0}
\definecolor{bicycle}{RGB}{255, 99, 71}
\definecolor{bus}{RGB}{255, 140, 0}
\definecolor{car}{RGB}{255, 69, 0}
\definecolor{const. veh.}{RGB}{233, 150, 70}
\definecolor{motorcycle}{RGB}{220, 20, 60}
\definecolor{pedestrian}{RGB}{255, 61, 99}
\definecolor{traffic cone}{RGB}{0, 0, 230}
\definecolor{trailer}{RGB}{47, 79, 79}
\definecolor{truck}{RGB}{112, 128, 144}
\definecolor{drive. suf.}{RGB}{0, 207, 191}
\definecolor{other flat}{RGB}{175,0,75}
\definecolor{sidewalk}{RGB}{75,0,75}
\definecolor{terrain}{RGB}{112,180,60}
\definecolor{manmade}{RGB}{222,184,135}
\definecolor{vegetation}{RGB}{0,175,0}

\usepackage[compact]{titlesec}
\titlespacing{\subsection}{0pt}{*0}{*0}
\titlespacing{\subsubsection}{0pt}{*0}{*0}

\title{SyntheOcc: Synthesize Geometric-Controlled \\Street View Images through 3D Semantic MPIs}

\author{%
Leheng Li$^{1}$  \quad Weichao Qiu$^{3}$  \quad Yingjie Cai$^{3}$  \quad Xu Yan$^{3}$   \quad Qing Lian$^{2}$ \\ \quad \textbf{Bingbing Liu}$^{3}$  \quad \textbf{Ying-Cong Chen}$^{1,2}$\thanks{Corresponding author.}  \quad  \vspace{8pt}\\
HKUST(GZ)\textsuperscript{1} \quad HKUST\textsuperscript{2} \quad HUAWEI Noah's Ark Lab\textsuperscript{3}
\\
\small{Project page: \texttt{\href{https://len-li.github.io/syntheocc-web/}{len-li.github.io/syntheocc-web}}}
}

\begin{document}

\maketitle

\begin{abstract}

The advancement of autonomous driving is increasingly reliant on high-quality annotated datasets, especially in the task of 3D occupancy prediction, where the occupancy labels require dense 3D annotation with significant human effort. In this paper, we propose \methodname, which denotes a diffusion model that \uline{Synthe}size photorealistic and geometric-controlled images by conditioning \uline{Occ}upancy labels in driving scenarios. This yields an unlimited amount of diverse, annotated, and controllable datasets for applications like training perception models and simulation. SyntheOcc addresses the critical challenge of how to efficiently encode 3D geometric information as conditional input to a 2D diffusion model. Our approach innovatively incorporates 3D semantic multi-plane images (MPIs) to provide comprehensive and spatially aligned 3D scene descriptions for conditioning. As a result, SyntheOcc can generate photorealistic multi-view images and videos that faithfully align with the given geometric labels (semantics in 3D voxel space). Extensive qualitative and quantitative evaluations of SyntheOcc on the nuScenes dataset prove its effectiveness in generating controllable occupancy datasets that serve as an effective data augmentation to perception models.

% By doing so, SyntheOcc can faithfully generate photorealistic multi-view images and videos that align with the given occupancy labels.

% Experimental results demonstrate that our generated dataset can serve as an effective data augmentation to perception models.

% This work notably propels the field of autonomous driving by effectively augmenting the training dataset used for advanced BEV perception techniques.

\end{abstract}
\section{Introduction}\label{sec:intro}

% key: occ is good, label occ is hard
% auto drive occ predict is data hungry
% diffusion is good at gen image
% bevgen bevctrl is not good
% how we did, our result

% Recent progress in diffusion models has greatly improved the integration of data synthesis with 2D control such as text prompts and image controls in ControlNet. However, achieving finer 3D geometric control in street view generation such as 3D semantic voxel, which is essential for 3D perception tasks, continues to be a challenge that has not yet been addressed. 

\begin{figure*}[!t]
    \centering
    \includegraphics[width = \textwidth, trim = 0 0 0 0, clip]{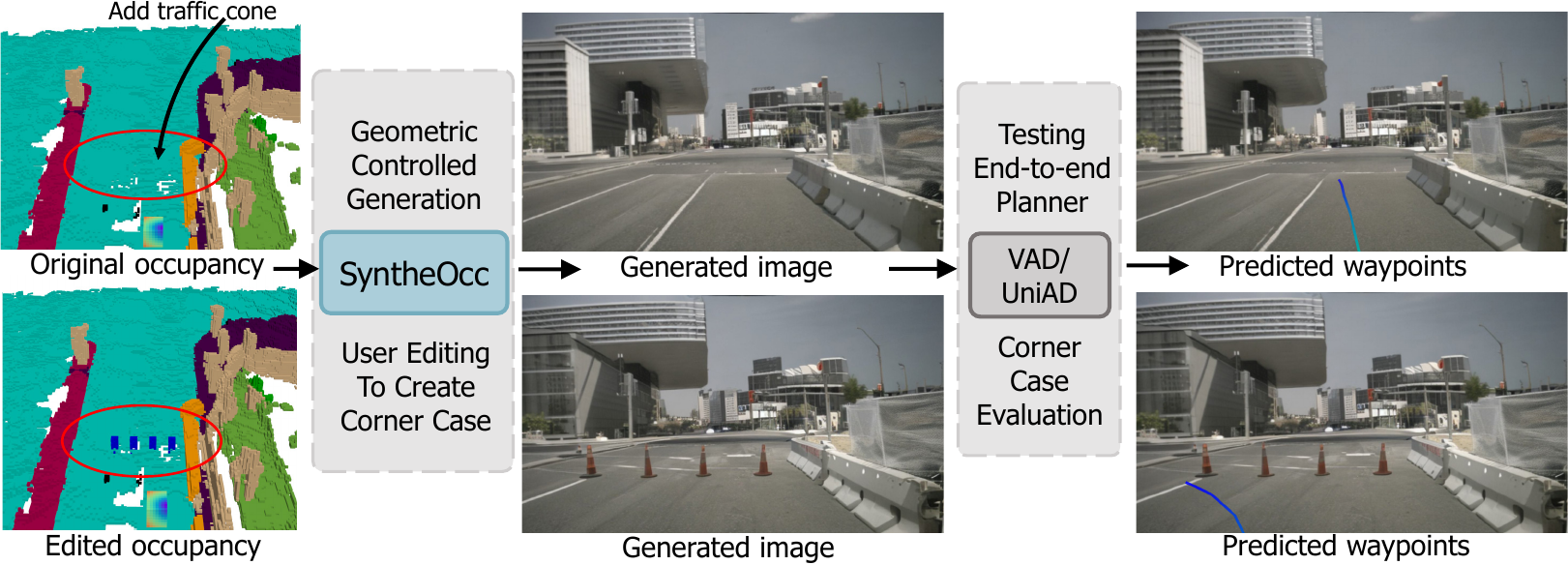}
    \caption{A showcase of application of \methodname. We enable geometric-controlled generation that conveys the user editing in 3D voxel space to generate realistic street view images. In this case, we create a rare scene that traffic cones block the way. This advancement facilitates the evaluation of autonomous systems, such as the end-to-end planner VAD~\cite{jiang2023vad}, in simulated corner case scenes.}
    \vspace{-1mm}
    \label{fig:e2etest}
\end{figure*}

With the rapid development of generative models, they have shown realistic image synthesis and diverse controllability. This progress has opened up new avenues for dataset generation in autonomous driving~\cite{gao2023magicdrive, swerdlow2024street, wen2023panacea, lift3D2023CVPR}. The task of dataset generation is usually modeled as controllable image generation, where the ground truth (\textit{e.g.} 3D Box) is employed to control the generation of new datasets in downstream tasks (\textit{e.g.} 3D detection). This approach helps to mitigate the data collection and annotation effort as it can generate labeled data for free. However, a novel task of vital importance, occupancy prediction~\cite{wang2023openoccupancy,tian2024occ3d}, poses new challenges for dataset generation compared with 3D detection. It requires finer and more nuanced geometry controllability, which refers to use the occupancy state and semantics of voxels in the whole 3D space to control the image generation. We argue that solving this problem not only allows us to synthesize occupancy datasets, but also empowers valuable applications such as editing geometry to generate rare data for corner case evaluation, as shown in Fig.~\ref{fig:e2etest}. In the following, we first illustrate why prior work struggles to achieve the above objective, and then demonstrate how we address these challenges.

% In the following, we first illustrate why prior work struggles to achieve the above objective. Then, we demonstrate how we address these challenges and further enable valuable applications such as generating rare data for corner case evaluation in Fig.~\ref{fig:e2etest}.

% This type of controllability enables valuable applications such as corner case generation in Fig.~\ref{fig:e2etest}.

In the area of diffusion models, several representative works have displayed high-quality image synthesis; however, they are constrained by limited 3D controllability: they are incapable of editing 3D voxels for precise control. For example, BEVGen~\cite{swerdlow2024street} generates street view images by conditioning BEV layouts using diffusion models. MagicDrive~\cite{gao2023magicdrive} extend BEVGen and additionally converts the 3D box parameters into text embedding through Fourier mapping that is similar to NeRF~\cite{mildenhall2020nerf}, and uses cross-attention to learn conditional generation. Although these methods achieve satisfactory results in image generation, their 3D controllability is inherently limited. These approaches are restricted to manipulating the scene in types of 3D boxes and BEV layouts, and hardly adapt to finer geometry control such as editing the shape of objects and scenes. Meanwhile, they usually convert conditional input into 1D embedding that aligns with prompt embedding, which is less effective in 3D-aware generation due to lack of spatial alignment with the generated images. This limitation hinders their utility in downstream applications, such as occupancy prediction and editing scene geometry to create long-tailed scenes, where granular volumetric control is paramount in both tasks.

ControlNet~\cite{zhang2023adding} and GLIGEN~\cite{li2023gligen} is another type of prominent method in the field of controllable image generation. These approaches exhibit several desirable attributes in terms of controllability. They leverage conditional images such as semantic masks for control, thereby offering a unified framework to manipulate both foreground and background. However, despite its precise spatial control, ControlNet does not align with our specific requirements. Their conditions of pixel-level images differ fundamentally from what we require in 3D contexts. Our experimental results also find that ControlNet struggles to handle overlapping objects with varying depths (see Fig.~\ref{fig:ctrl_compare}~(a)), as it only utilizes an ambiguous 2D semantic map as conditional input. As a result, it is non-trivial to extend the ControlNet framework and convey their desirable attributes for 3D conditioning.

To address the above challenges, we propose an innovative representation, 3D semantic multi-plane images (MPIs), which contribute to image generation with finer geometric control. In detail, we employ multi-plane images~\cite{zhou2018stereo} to represent the occupancy, where each plane represents a slice of semantic label at a specific depth. Our 3D semantic MPIs not only preserve accurate and authentic 3D information, but also keep pixel-wise alignment with the generated images. We additionally introduce the MPI encoder to encode features, and the reweighing methods to ease the training with long-tailed cases. As a collection, our framework enables 3D geometry and semantic control for image generation and further facilitates corner case evaluation as depicted in Fig.~\ref{fig:e2etest}. Finally, experimental results demonstrate that our synthetic data achieve better recognizability, and are effective in improving the perception model on occupancy prediction. In summary, our contributions include:
% Our contributions are summarized as follows:

    \vspace{-1mm}

\begin{itemize}
    \item We present \methodname, a novel image generation framework to attain finer and precise 3D geometric control, thereby unlocking a spectrum of applications such as 3D editing, dataset generation, and long-tailed scene generation. 
    %  long-tailed scene generation.  corner case evaluation  (Fig.~\ref{fig:e2etest})

    \item Incorporating the proposed 3D semantic MPI, MPI encoder, and reweighing strategy, we deliver a substantial advancement in image quality and recognizability over prior works.
    
    \item Our extensive experimental results demonstrate that our synthetic data yields an effective data augmentation in the realm of 3D occupancy prediction.
\end{itemize}

    \vspace{-1mm}

% Compared with previous methods~\cite{gao2023magicdrive,swerdlow2024street,wen2023panacea} restricted to editing the 3D boxes, 

\vspace{-2mm}

% key: occ is good, label occ is hard
% auto drive occ predict is data hungry
% diffusion is good at gen image
% bevgen bevctrl is not good
% how we did, our result

\section{Related Work}\label{sec:related}

    \vspace{-2mm}

% 3D Occupancy Prediction
% Diffusion-based Generative Models 
% Dataset generation in autonomous driving

\subsection{3D Occupancy Prediction}
The task of 3D occupancy prediction aims to predict the occupancy status of each voxel in 3D space, as well as its semantic label if occupied. Compared with previous perception methods like 3D object detection, occupancy prediction offers a more detailed and nuanced understanding of the environment, as it provides finer geometric details, is capable of handling general, out-of-vocabulary objects, and finally, enriches the planning stack with comprehensive 3D information. Early methods exploited LiDAR as inputs to complete the 3D occupancy of the entire 3D scene~\cite{yan2021sparse,mei2023ssc}. Recent methods began to explore the more challenging vision-based 3D occupancy prediction~\cite{wang2023openoccupancy,tian2024occ3d,tong2023scene,wei2023surroundocc}. By predicting the geometric and semantic properties of both dynamic and static elements, 3D occupancy prediction offers a more comprehensive understanding of the surrounding environment.

    \vspace{-1mm}

\subsection{Diffusion-based Image Generation}
    \vspace{-1mm}

Recent advancements in diffusion models (DMs) have achieved remarkable progress in image generation. In particular, Stable Diffusion (SD)~\cite{rombach2022high} employs DMs within the latent space of autoencoders, striking a balance between computational efficiency and high image quality. Beyond text control, there is also the introduction of additional control signals. A noteworthy work is ControlNet~\cite{zhang2023adding}, which incorporates a trainable copy of the SD encoder to extract the feature of conditional images and adds it to the UNet feature. It significantly enhances the controllability and unlocking pathways for advanced applications. We refer readers to recent survey~\cite{yang2023diffusion} for more details.

% Furthermore, some studies concentrate on generating multi-view images. 

\subsection{Image Generation in Autonomous Driving}
    \vspace{-1mm}
As training neural networks relies heavily on labeled
data, numerous studies are delving into dataset generation to boost training. Lift3D~\cite{lift3D2023CVPR} designs generative NeRF to synthesize labeled datasets for 3D detection for the first time. Several other works employ BEV layouts to synthesize image data, proving beneficial for perception models. For example, BEVGen~\cite{swerdlow2024street} conditions BEV layouts to generate multi-view street images, while BEVControl~\cite{yang2023bevcontrol} separately generates foregrounds and backgrounds from BEV layouts. MagicDrive~\cite{gao2023magicdrive} generates images with 3D geometry controls by independently encoding objects and maps through a text encoder or map encoder. Compared with MagicDrive, our geometry control is characterized by a more detailed and lossless representation of 3D scenes for control, which poses significant challenges than projected layout or box embedding.

Recently, DriveDreamer~\cite{wang2023drivedreamer}, DrivingDiffusion~\cite{li2023drivingdiffusion}, Drive-WM~\cite{wang2023driving} and Panacea~\cite{wen2023panacea} use a ControlNet framework, which involves projecting bounding boxes and road maps onto 2D FoV images as a conditioning input. This approach has proven to be effective for geometric control. However, it is limited in that it only achieves alignment at the 2D-pixel level. Consequently, this method falls short in capturing the depth hierarchy and fails to account for the occlusion relationships present in the 3D real world. Besides, adding a depth channel like Panacea~\cite{wen2023panacea} may address the limitations of depth order, but it discards the occluded part and only contains partial observation. UrbanGiraffe~\cite{yang2023urbangiraffe} train a generative NeRF to perform image generation. WoVoGen~\cite{lu2023wovogen} creates a 4D world volume feature using occupancy to guide the generation, but seems to rely on object mask guidance.

As described above, most of the prior work is restricted by only modeling a projected primitive of 3D boxes and road maps as conditions. They suffer from ill-posed un-projection ambiguity. In contrast, we model 3D occupancy labels as conditions, as they provide finer geometric details and semantic information. However, designing an input representation of 3D occupancy labels into a 2D diffusion model is challenging. In this paper, we propose a novel representation: 3D semantic Multi-Plane Images (MPIs) as conditional inputs, which not only provide spatial alignment that improves visual consistency, but also encode comprehensive 3D geometric information including occluded parts.

% A naive baseline can be using projected occupancy labels as conditions and training a ControlNet.

% There are also recent work~\cite{lu2023wovogen} that use occupancy annotation to generate images.

\section{Method}\label{sec:method}

\begin{figure*}[!t]
    \centering
    \includegraphics[width = \textwidth, trim = 0 0 0 0, clip]{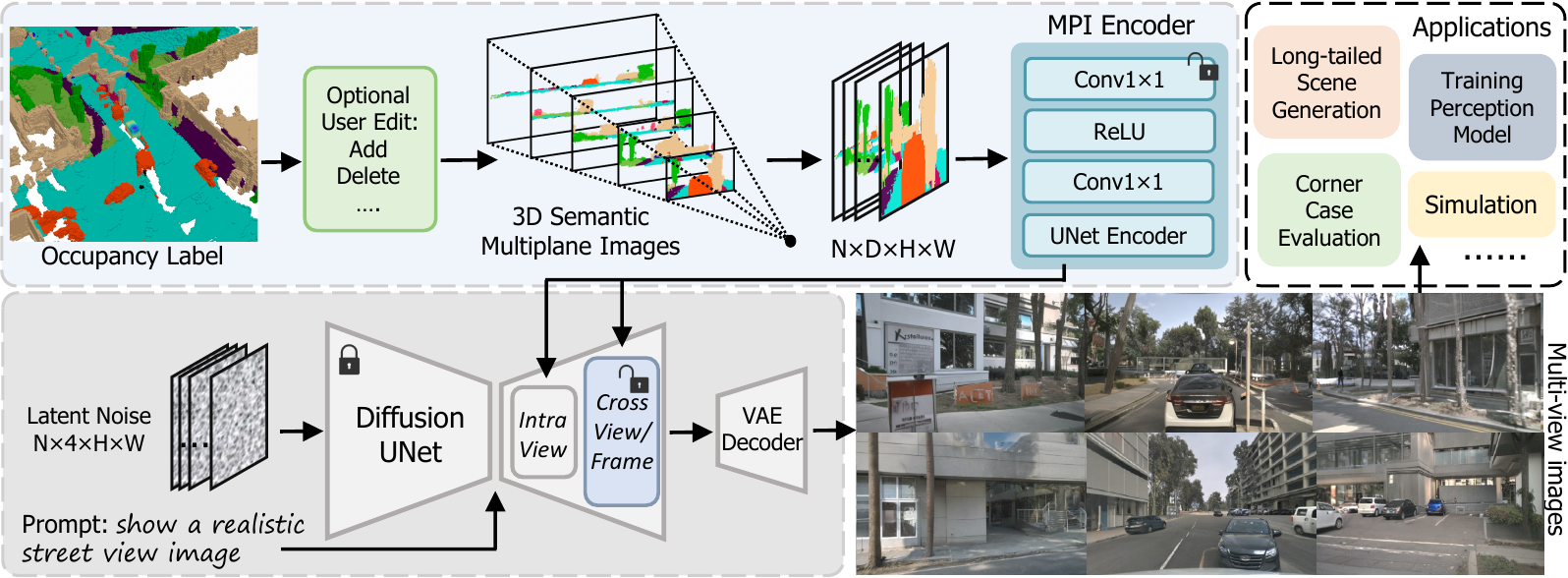}
    \caption{The overall architecture of \methodname. We achieve 3D geometric control in image generation by utilizing our proposed 3D semantic multiplane images to encode scene occupancy. In our framework, we can edit the occupied state and semantics of every voxel in 3D space to control the image generation, thereby opening up a wide spectrum of applications as shown in the top right. }
    % \vspace{-1mm}
    \label{fig:overview}
\end{figure*}

% We additionally propose MPI encoder to extract the 3D feature as the condition into the UNet. 

% We can edit the occupied state and semantics of every voxel in 3D space to control the image generation, thereby opening up a wide spectrum of applications as shown in the top right.

    \vspace{-2mm}

\myparagraph{Overview} The overview of our method is depicted in Fig.~\ref{fig:overview}. Built upon the SD pipeline, we aim to perform geometry-controlled image generation by conditioning on 3D geometry labels with semantics (occupancy labels). One requirement is that the images should faithfully align with the given label. This task is more challenging than conditioned on 3D box due to the sparse and irregular nature of occupancy. We first discuss how to efficiently represent occupancy in Sec.~\ref{sec:repocc}, followed by our designed MPI encoder to enhance generation quality in Sec.~\ref{sec:occenc}, and reweighing strategy to handle the long-tailed depth and category in Sec.~\ref{sec:rew}.

% Local Condition Aligns Better than Global Condition

% \subsection{Type of Condition: Global or Local?}\label{sec:typecondi}
\subsection{Representation of Condition: Local Control Aligns Better than Global Control}\label{sec:typecondi}
One of the key challenges is how to represent our conditional occupancy input. A straightforward method~\cite{gao2023magicdrive,chen2023integrating} is to convert the 3D occupancy voxel to 1D global embedding that is similar to text embedding, and then use cross-attention to learn controllable generation. However, these global methods can be less effective when dealing with dense or irregular data due to the following reasons: \textbf{(i)} They perform controllable generation through hard encoding the spatial relationship between 1D global embedding and 2D UNet features. \textbf{(ii)} Ignore the underlying geometry alignment between the conditional input and the generated image. In contrast, local methods like ControlNet, directly add spatial features to the UNet features, providing 2D local control with pixel-level spatial alignment. They are better than the global method (see Tab.~\ref{tab:base_main}), but suffer from 3D ambiguity (see Fig.~\ref{fig:ctrl_compare}~(a)). Consequently, this comparison motivates us to seek a more compact and efficient manner to encode and condition our 3D occupancy labels.

% In this manner, any dimension and type of conditions can be injected into the diffusion model. 
% Losing geometric details. 

% Consequently, this limitation motivates us to extend the ControlNet framework to 3D conditioning. 

% Consequently, this comparison motivates us to seek a more compact and efficient representation to encode 3D occupancy labels.
% It is non-trivial to extend ControlNet to 3d

\subsection{Represent Occupancy as 3D Semantic Multiplane Images}\label{sec:repocc}
% \subsection{Multiplane Images Encoding for Occupancy: a unified approach}

It is non-trivial to design a 3D representation for conditioning. To efficiently store both the semantic and geometric information of the irregular occupancy input, we propose to use multiplane images (MPIs)~\cite{zhou2018stereo} as representation. An MPI is composed of a series of fronto-parallel RGBA layers within the frustum of the source camera with a specific viewpoint. These planes are arranged at varying depths, from $d_{min}$ to $d_{max}$, starting from the nearest to the farthest. Each layer of these images contains both an RGB image and an alpha map, which collectively capture the visual and geometric details of the scene at the respective depth. In our work, instead of storing RGB value and alpha map in the original MPI, we store our 3D semantic labels. Each layer of MPI represents the semantic index at the corresponding depth. We display the colored MPI in the top row of Fig.~\ref{fig:overview} for visual clarity, but we actually use the integer index for learning. We obtain our 3D semantic MPI by:
\begin{gather}
P_{l}=(u\times  d_{l}, ~v\times d_{l}, ~d_{l})^{T},~d_{l}=d_{min}+(d_{max}-d_{min})\times l / D, \\
\text{\mpiname}_{n,l} = \texttt{Interpolate}(\text{\occname},~\mathbf{T_n}\cdot \mathbf{K^{-1}_n}\cdot P_{l}), \\
% \text{\mpiname} = \texttt{Concat}(\text{\mpiname}_{0},\dots,\text{\mpiname}_{l},\dots,\text{\mpiname}_{D}),
\text{\mpiname} = \texttt{Concatenate}(\text{\mpiname}_{i,j}),~i\in (0,N),~j\in (0,D),
\label{equ:mtp}
\end{gather}
% \begin{gather}
% P(u,v,l)=(u\times  d_{l}, ~v\times d_{l}, ~d_{l})^{T},~d_{l}=d_{min}+(d_{max}-d_{min})\times l / D, \\
% \text{\mpiname}(n,l,v,u) = \texttt{Interpolate}(\text{\occname},~\mathbf{M_n}\cdot \mathbf{K^{-1}_n}\cdot P(u,v,l)), \\
% % \text{\mpiname} = \texttt{Concat}(\text{\mpiname}_{0},\dots,\text{\mpiname}_{l},\dots,\text{\mpiname}_{D}),
% \label{equ:mtp}
% \end{gather}
% \begin{equation}
% \begin{split}
% x=\frac{(u-c_U)\times d_{l}}{f_U} , y=\frac{(v-c_V)\times d_{l}}{f_V}, z=d_{l}, \\
% \text{\mpiname}_{l} = \texttt{Interp}(\text{\occname}, (x,y,z)), 
% \label{equ:mtp}
% \end{split}
% \end{equation}
where $(u, v)$ is a pixel coordinate in image space, $d_{l}$ is depth value of the $l^{th}$ layer, $n$ denotes the $n^{th}$ camera view. This equation implies we first back project points $P$ in camera frustum space $(u,v,d)$ to Euclid space $(x,y,z)$ by multiplying inverse intrinsic $\mathbf{K^{-1}}$. Then we use transformation matrix $\mathbf{T}$ to map points from camera coordinates to occupancy coordinates. We then use the point coordinates to interpolate the nearest semantic index from the dense occupancy voxel to form a slice of MPI. Finally, we concatenate all slices to form $\text{\mpiname} \in \mathbb{R}^{N \times D \times H \times W}$, where $D$ is the number of layers that is set at 256, $N$ is the number of camera views in the case of batch size = 1.

% is the multiplication of batch size, view numbers, and optionally frame numbers.

% The dimension of an $\text{\mpiname}$ is typically $D\times H\times W$, where $D$ is the number of layers that is set at 256.

By representing occupancy as 3D semantic MPI, every pixel in MPI contains geometry and semantic information with implicit depth, seamlessly integrating occluded elements, and ensuring a precise spatial alignment with the generated images.

% We display the 3D geometry and semantic control in Fig.~\ref{fig:edit}.

% This innovative method affords us 3D geometry and semantic control as shown in Fig.~\ref{fig:edit}.

% , and provides better efficacy than previous methods that will be discussed in Sec.~\ref{sec:experiments}. 

% We store only the semantic index.
% In this manner, since we store geometric and semantic information in a unified way, we achieve satisfactory spatial alignment results.

% As shown in Fig.1, we cast 256 layers of multiplane images.

\subsection{3D Semantic MPI Encoder}\label{sec:occenc}

% \subsection{Occupancy Encoder}\label{sec:occenc}

% In order to achieve a spatially aligned generation effect, we design a novel type of MPI encoder. 

% Inspired by the architecture of ControlNet~\cite{zhang2023adding}

% In order to transfer the locality property of the ControlNet framework to 3D,

% In order to achieve a spatially aligned condition,

To enable local control with spatially aligned conditions, we develop a simple but effective MPI encoder that aligns the 3D multi-plane feature to the latent space of the diffusion model. The purpose of the MPI encoder is to obtain features from multi-plane images to perform 3D-aware image synthesis. Unlike the original ControlNet which downsampling conditional input through 3×3 convolutions with padding, we design a 1×1 convolutional encoder without downsampling to encode features. In detail, the 3D multiplane features which have the sample resolution with latent features, are transformed by a 1×1 convolution layer and ReLU activation~\cite{agarap2018deep} in the MPI encoder. 

% If we use a 3×3 convolution, it will be conducted in a camera frustum space rather than Euclid space.

\begin{figure*}[!t]
    \centering
    \includegraphics[width = \textwidth, trim = 0 0 0 0, clip]{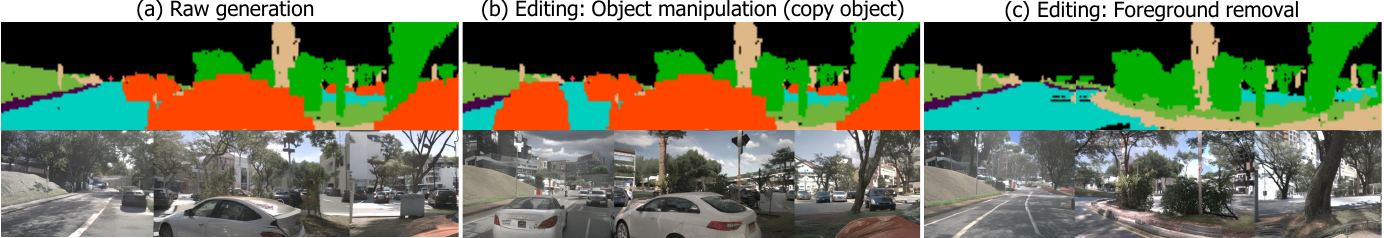}
    \caption{Visualizations of geometric controlled generation. \textbf{Top row}: Fusion of 3D semantic MPI. \textbf{Bottom row}: our generation concatenated from neighboring views.}
    \vspace{-3mm}
    \label{fig:edit}
\end{figure*}

After obtaining the multi-scale feature after the MPI encoder, we add the feature to the decoder of diffusion UNet to provide spatial features. Experimental results in Tab.~\ref{tab:ablate} will show that our 1×1 conv in MPI encoder is more effective than 3×3 conv, as the 1×1 conv with receptive field $=1$ provides a spatial align feature to the latent feature in the diffusion UNet. In contrast, 3×3 conv is conducted in a camera frustum space rather than Euclid space, making an imprecise correspondence between 3D multiplane features and 2D image features. Moreover, using 3×3 conv to process 3D semantic MPI will introduce a large computational burden as the channel number increases from 3 channels of RGB to 256 planes. We display our 3D geometry and semantic control property in Fig.~\ref{fig:edit}.

% This innovative method affords us 3D geometry and semantic control as shown in Fig.~\ref{fig:edit}.

% : $U_{out}=U_{in}+Encoder(\text{\mpiname})$

% 3×3 convolution with receptive field $>1$ will destroys the correspondence between 3D multiplane features and 2D image features, leading to a decrease in performance.

In summary, we chose MPIs as the representation because they \textbf{(i)} Incorporate lossless 3D information, including scene geometry rather than 2.5D depth. \textbf{(ii)} Provide spatially aligned conditional features that naturally extend the ControlNet framework from image level to 3D level. \textbf{(iii)} Capable of representing geometry and semantics including occluded elements.

% and successfully convey the strength of ControlNet architecture

% Although we receive downsampled MPIs, the limited resolution of the occupancy grid does not hinder learning.

    % \vspace{-1mm}

% \subsection{Cross-View Attention}\label{sec:attention}
\subsection{Cross-View and Cross-Frame Attention}\label{sec:attention}

    \vspace{-1mm}

The sensor arrangement in a self-driving car usually requires a full surround view of cameras to capture the entire 360-degree environment. To effectively simulate the multi-view and subsequent multi-frame generation, zero-initialized~\cite{zhang2023adding} cross-view and cross-frame attention are integrated into the diffusion model to maintain consistency between views and frames. Following prior work~\cite{wu2023tune,gao2023magicdrive,wen2023panacea,wang2023driving}, each cross-view attention allows the target view to access information from its neighboring left and right views, thus training cross-view attention using multi-view consistent images will enforce it to generate the same instance in the overlapping region of multi-view cameras. 
\begin{align}
    \texttt{Attention}(Q, K, V) = \texttt{softmax}(\begin{matrix}\frac{QK^{T}}{\sqrt{d}}\end{matrix}) \cdot V\text{,}
    \label{equ:attn} \\
    % {h}_{cross\_view}^{v} = {h}_{in}^{v}  + \sum_{i\in\{l,r\}}\texttt{Attention}(Q_{in}, K_{i}, V_{i})\text{,} \\
    {h}_{out} = {h}_{in}  +  {\textstyle \sum_{i\in\{l,r\}}^{}} \texttt{Attention}(Q_{in}, K_{i}, V_{i})\text{,}
    \label{equ:skip-attn}
\end{align}
where $l$, and $r$ is the camera view of left and right. $Q_{in}$ and ${h}_{in}$ denotes the query and the hidden state of input view. Similarly, we add cross-frame attention that attend previous frame and future frame to enable video generation. In this case, we use the same formulation while $i\in\{f,h\}$, where $f$ and $h$ is the camera view of future and history frames. 

% The generated video is provided in the appendix.

    \vspace{-1mm}

\subsection{Importance Reweighing}\label{sec:rew}

    \vspace{-1mm}

\begin{wraptable}{r}{6cm}
% \arrayrulecolor[HTML]{DB5800}
\centering
\resizebox{1.0\textwidth}{!}{
    \begin{minipage}[t][][b]{1\textwidth}
        \centering
        \includegraphics[width=1.0\textwidth]{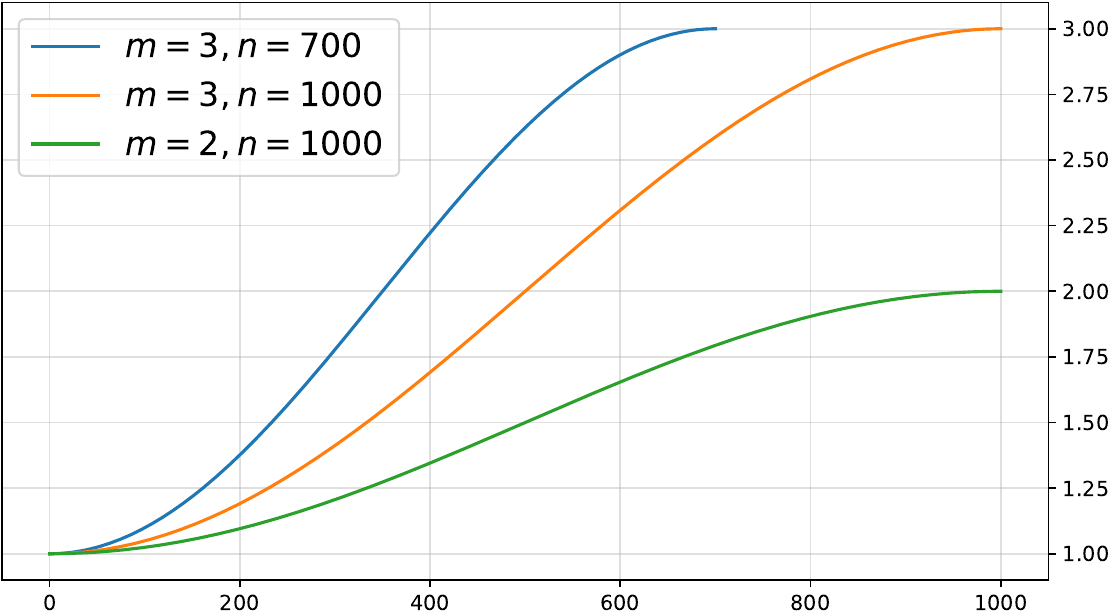}
        \vspace{-0.1cm}
        \captionof{figure}{Visualizations of the reweighing function in Eq.~\ref{equ:reweight}.}\label{fig:rew_vis}
    \end{minipage}
}
\vspace{-0.6cm}
\end{wraptable}

To deal with the extreme imbalance problem between foreground, background, and object categories, and also to ease the training, we propose three types of reweighting methods to improve the generation quality of foreground objects.

    \vspace{-2mm}

\myparagraph{Progressive Foreground Enhancement} To mitigate the complexity of the learning task, we propose a progressive reweighting method that incrementally enhances the loss associated with the foreground regions (based on semantic class) as the training progresses. The detailed formulation is:  
\begin{equation}
    w_{}(x,m,n) = \frac{(m-1)}{2} \cdot ( 1 + \cos(\frac{x}{n_{}} \cdot \pi + \pi)) + 1,
\label{equ:reweight}
\end{equation}
where $x$ is the current training step, $m$ is the maximum value of weights that set at $2$, and $n$ is the total training steps. This approach is engineered to facilitate a learning trajectory that progresses from simplicity to complexity, thereby aiding in the convergence of the model. This curve can be interpreted as a cosine annealing but inverted to amplify the importance of the foreground region.

% , where $n=1000$

\myparagraph{Depth-aware Foreground Reweighing} In the meantime, we acknowledge the learning difficulty in different depth places in 3D scenes. Following GeoDiffusion~\cite{chen2023integrating}, we perform depth reweighing to foreground objects by adaptively assigning higher weights to farther foreground areas. This enables the model to focus more thoroughly on hard examples with depth-aware importance reweighting. Instead of using their exponential function to increase weights, we use our designed cosine function Eq.~\ref{equ:reweight} for stability. Here $x$ is the input depth value, and $n$ is the maximum depth that set at $50$.

\myparagraph{CBGS Sampling} To deal with the class imbalance problem in driving scenarios, where certain object categories appear infrequently, we employ the Class-Balanced Grouping and Sampling (CBGS)~\cite{zhu2019class} to better handle the long-tailed classes. CBGS addresses the challenge of class imbalance by grouping and re-sampling training data to ensure each group has a balanced distribution of sample frequency across different object categories. This method reduces the bias towards more frequent classes and enables better generalization to rare scenarios.

\begin{figure*}[!t]
    \centering
    \includegraphics[width = \textwidth, trim = 0 0 0 0, clip]{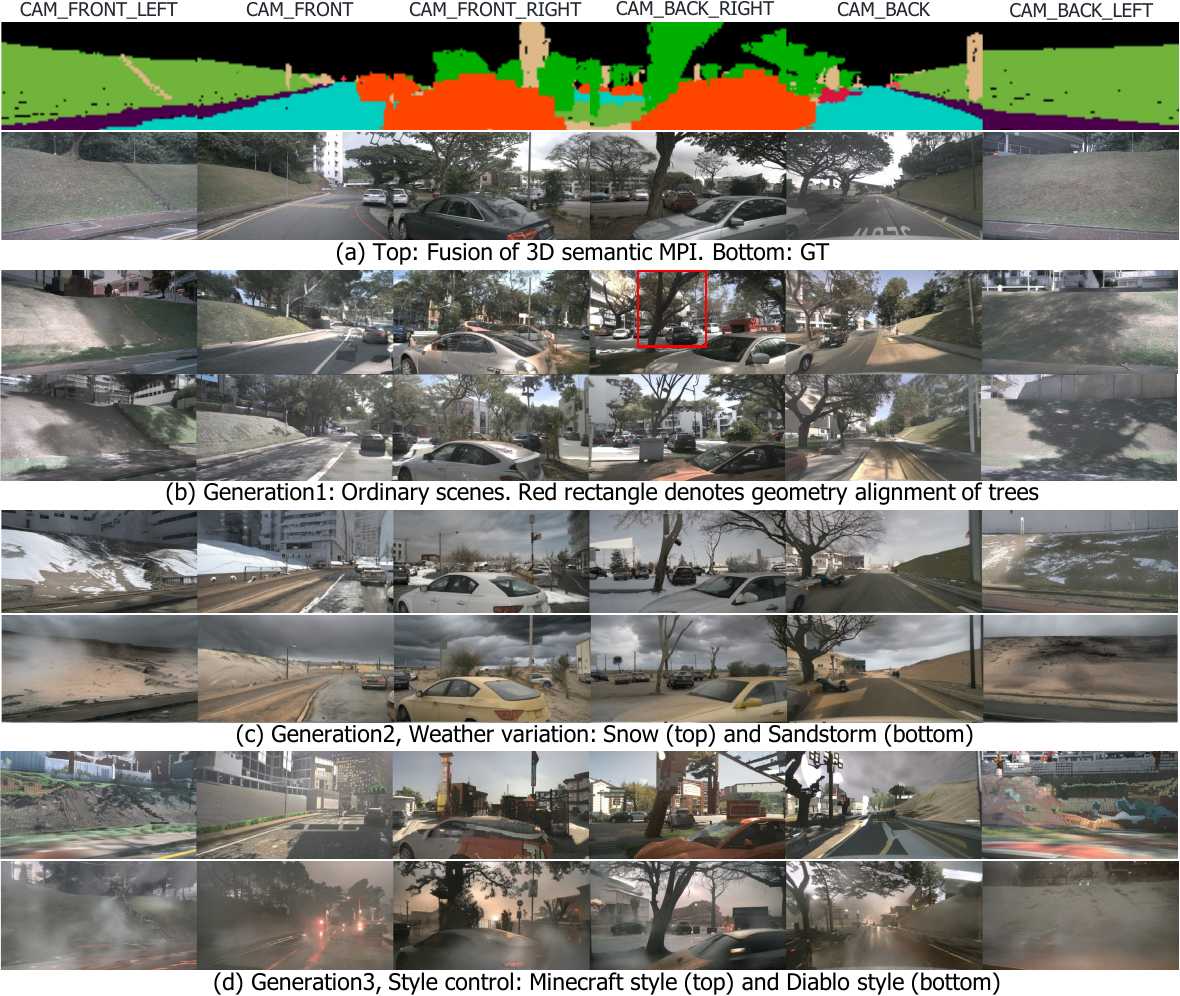}
    \caption{Visualizations of generated multi-view images. The generation conditions (occupancy labels) are from nuScenes validation set. We highlight that \textbf{(i)} Geometry alignment of trees in red rectangle in (b). \textbf{(ii)} Use text prompt to control high-level appearance in (c,d).}
    % \vspace{-1mm}
    \label{fig:infer}
\end{figure*}

\subsection{Model Training}\label{sec:modeltrain}

To ease the training of the MPI encoder and added attention module, we use a two stage training pipeline. We first train MPI encoder and cross-view attention in a multi-view image generation setting. Then we train cross-frame attention and freeze other components in a video generation setting.

% \textbf{}
\myparagraph{Objective Function} Our final objective function can be formulated as a standard denoising objective with reweighing:
\begin{equation}
    \mathcal{L}_{\rm} = \mathbb{E}_{\mathcal{E}(x),\epsilon,t}\|\epsilon - \epsilon_\theta(z_t, t, \tau_\theta(y))\|^2\odot w,
    \label{equ:objective}
\end{equation}
where $w$ is the multiplication of progressive reweighing and depth-aware reweighing.

% \subsection{Model Training}

% \subsubsection{Classifier-free Guidance}

% \subsubsection{Training Objective and Augmentation}

\section{Experiments}\label{sec:experiments}

    \vspace{-2mm}

\subsection{Dataset and Setups}
We conduct our experiments on the nuScenes dataset~\cite{nuscenes}, which is collected using 6 surrounded-view cameras that cover the full 360° field of view around the ego-vehicle. It contains 700 scenes for training and 150 scenes for validation. We resize the original image from 1600 × 900 to 800 × 448 for training. In our work, we use the occupancy label with a resolution of $0.2m$ from OpenOccupancy~\cite{wang2023openoccupancy} as condition input, while the benchmark of occupancy prediction uses a resolution of $0.4m$ from Occ3D~\cite{tian2024occ3d} dataset for its popularity.

\begin{table*}
	\setlength{\tabcolsep}{0.0035\linewidth}
	\newcommand{\classfreq}[1]{{~\tiny(\semkitfreq{#1}\%)}}  %
	\centering
   \resizebox{1\linewidth}{!}{
	\begin{tabular}{l|c c | c |c  c c c c c c c c c c c c c c c c}
 
		\toprule
		\textbf{Method}
		& \makecell[c]{\textbf{Train} }
		& \makecell[c]{\textbf{Val} }
            & \makecell[c]{\textbf{mIoU}}
		& \rotatebox{90}{\textcolor{barrier}{$\blacksquare$} barrier} 
		& \rotatebox{90}{\textcolor{bicycle}{$\blacksquare$} bicycle}
		& \rotatebox{90}{\textcolor{bus}{$\blacksquare$} bus} 
		& \rotatebox{90}{\textcolor{car}{$\blacksquare$} car} 
		& \rotatebox{90}{\textcolor{const. veh.}{$\blacksquare$} cons. veh.} 
		& \rotatebox{90}{\textcolor{motorcycle}{$\blacksquare$} moto.} 
		& \rotatebox{90}{\textcolor{pedestrian}{$\blacksquare$} pedes.} 
		& \rotatebox{90}{\textcolor{traffic cone}{$\blacksquare$} traf. cone} 
		& \rotatebox{90}{\textcolor{trailer}{$\blacksquare$} trailer} 
		& \rotatebox{90}{\textcolor{truck}{$\blacksquare$} truck} 
		& \rotatebox{90}{\textcolor{drive. suf.}{$\blacksquare$} drive. suf.} 
		& \rotatebox{90}{\textcolor{other flat}{$\blacksquare$} other flat} 
		& \rotatebox{90}{\textcolor{sidewalk}{$\blacksquare$} sidewalk} 
		& \rotatebox{90}{\textcolor{terrain}{$\blacksquare$} terrain} 
		& \rotatebox{90}{\textcolor{manmade}{$\blacksquare$} manmade} 
		& \rotatebox{90}{\textcolor{vegetation}{$\blacksquare$} vegetation} \\
		% & mIoU\\
		\midrule

            \textcolor{gray}{Oracle (FB-Occ~\cite{li2023fb})} & \textcolor{gray}{Real} &  \textcolor{gray}{Real} & \textcolor{gray}{39.3}  & \textcolor{gray}{45.4}  & \textcolor{gray}{28.2} &  \textcolor{gray}{44.1} &  \textcolor{gray}{49.4} & \textcolor{gray}{25.9}  & \textcolor{gray}{28.8}  & \textcolor{gray}{28.0}  &  \textcolor{gray}{27.7} & \textcolor{gray}{32.4} & \textcolor{gray}{37.3} & \textcolor{gray}{80.4} & \textcolor{gray}{42.2} & \textcolor{gray}{49.9}  & \textcolor{gray}{55.2} & \textcolor{gray}{42.0}  & \textcolor{gray}{37.7}  \\
            
            \methodname-Aug & Real+Gen &  Real & 40.3 &45.4 & 27.2  & 46.6 &  49.5 &  26.4 & 27.8  & 28.4  & 29.4  &  34.0 & 37.2 & 81.3 & 46.0 & 52.4 & 56.5  & 43.3 & 38.9    \\
            % \methodname-Aug & Real+Gen &  Real & 40.3 &45.4 & 27.2  & 46.6 &  49.5 &  26.4 & 27.8  & 28.4  & 29.4  &  34.0 & 37.2 & 81.3 & 46.0 & 52.4 & 56.5  & 43.3 & 38.9    \\
            
            \midrule
            % HardEncoding & Real &  Gen  & 13.4  & 0.7  & 0.0 &  11.8 & 32.4 & 0.0  & 6.6 & 2.8  &  0.3 & 2.6 & 19.6 & 60.1 & 12.1 & 26.2  & 23.4 & 15.5  & 12.8 \\

            MagicDrive & Real &  Gen  & 13.4  & 0.7  & 0.0 &  11.8 & 32.4 & 0.0  & 6.6 & 2.8  &  0.3 & 2.6 & 19.6 & 60.1 & 12.1 & 26.2  & 23.4 & 15.5  & 12.8 \\
            
            ControlNet & Real &  Gen & 17.3 & 17.7  & 0.2 & 13.6 &  21.0 &  0.6 & 0.8  & 8.6  & 10.4  &  6.9 & 11.9 & 67.4 & 18.8 & 36.4 & 36.9  & 20.8 & 22.4    \\
            ControlNet+depth & Real &  Gen  & 17.5  & 19.3 & 0.3 &  14.0 &  23.7 & 1.0  & 0.6  & 9.2  &  9.2 & 5.7 & 12.1 & 68.8 & 19.2 & 36.0  & 35.3 & 19.8  & 22.8  \\
            \methodname-Gen & Real &  Gen  & 25.5  & 32.6 & 13.8 &  27.7 &  33.4 & 7.5  & 6.5  & 15.7  &  16.5 & 16.5 & 25.6 & 74.3 & 24.5 & 39.4  & 40.5 & 28.6  & 28.8  \\

		\bottomrule
	\end{tabular}}\\

	\caption{Downstream evaluation on the \textbf{nuScenes-Occupancy} validation set. Based on the used train and val data, two types of settings are reported. The first is to use generated training set to augment the real training set, and evaluate on the real validation set, denoted as Aug. The second is to use pretrained models trained on the real training datasets to test on the generated validation set, denoted as Gen. }
	\label{tab:base_main}
\end{table*}

    \vspace{-1mm}

\myparagraph{Networks} We use Stable Diffusion~\cite{rombach2022high} v2.1 checkpoint as initialization and only train occupancy encoder, cross-view attention. We additionally add cross-frame attention if in video experiments. We adopt FB-Occ~\cite{li2023fb} as the target model for occupancy prediction for its SOTA performance in this task. The pretrained checkpoint of the network is obtained from their official repository. Since FB-Occ predicts occupancy using only single frame images, we thus train SyntheOcc without cross-frame attention in related experiments. For video generation, we provide experimental results in appendix.

\myparagraph{Metrics} We use Frechet Inception Distance (FID)~\cite{heusel2017gans} to measure the perceptual quality of generated images, and use mIoU to measure the precision of occupancy prediction.
    \vspace{-1mm}

\myparagraph{Hyperparameters} We set $D=256$, $d_{min}=0$ and $d_{max}=50$. The depth resolution of MPI is thus higher than occupancy voxel. We train our model in 6 epochs with batch size $=8$. The learning rate is set at $2e^{-5}$. The training phase takes around 1 day using 8 NVIDIA A100 80G GPUs. We use UniPC scheduler ~\cite{zhao2023unipc} with the classifier-free guidance (CFG)~\cite{ho2022cfg} that is set as 7.0. During inference, we use 20 denoising steps for dataset generation. 

% , and 40 denoising steps for visualization

    \vspace{-1mm}

 % (Fig.~\ref{fig:ctrl_compare}~(a0))
 
\myparagraph{Baselines} We compare our method with prior methods in Tab.~\ref{tab:base_main}. ControlNet denotes we train a ControlNet using an RGB semantic mask as the condition. ControlNet+depth denotes we add a depth channel after the semantic mask to provide 2.5D depth information. The depth map rendered by occupancy is normalized to [0-255] to accommodate the RGB value. The ControlNet+depth can be regarded as a degradation of \methodnamesmall which is reduced to a single plane. Then we evaluate MagicDrive since it is the only open-sourced method in this area. MagicDrive separately encodes foreground and background using prompt and BEV layout. Furthermore, we evaluate the image quality (FID~\cite{heusel2017gans}) of our method in Tab.~\ref{tab:comparefid}. Compared with prior methods, we use a unified 3D representation that seamlessly handles foreground and background, surpassing them by a large margin.

 % (may be due to imprecise alignment)
 
% Compared with them, we use a unified 3D representation that seamlessly handles foreground and background,  

\subsection{Qualitative Results}
    \vspace{-1mm}

\begin{figure*}[!t]
    \centering
    \includegraphics[width = \textwidth, trim = 0 0 0 0, clip]{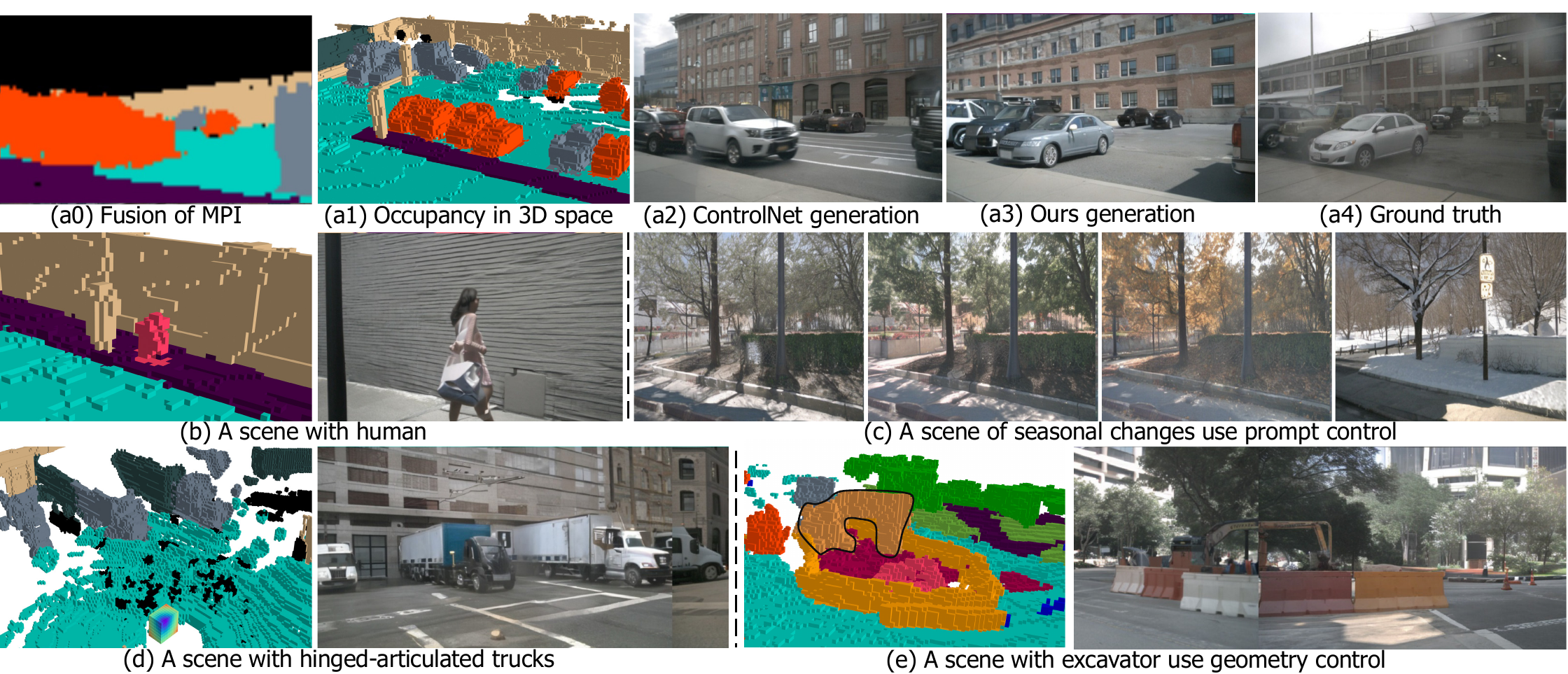}
    \caption{\textbf{Top row}: Comparison with ControlNet. We achieve a precise alignment between conditional labels and synthesized images, while ControlNet generates objects with incorrect pose due to ambiguous 2D condition. \textbf{Mid and Bottom row}: Visualizations of geometry-controlled image generation. We can faithfully generate objects with the desired topology in a specific 3D position.}
    % \caption{Left: Comparison of our method with baseline. Right: Visualizations of generated images.}
    % \vspace{-1mm}
    \label{fig:ctrl_compare}
\end{figure*}

\myparagraph{High-level Control using Prompt} In Fig.~\ref{fig:infer}~(c,d) and Fig.~\ref{fig:ctrl_compare}~(c), we demonstrate the capability to employ user-defined prompts to generate images with specific weather conditions and high-level style. Although the nuScenes dataset doesn't contain rare weather images like snow and sandstorms, our method successfully conveys prior knowledge pretrained from stable diffusion to our scenes. Compared with visualization results in prior work like Fig. 8 of MagicDrive, our method shows better alignment with the text prompt, demonstrating the cross-domain generalization ability of our method.

    \vspace{-1mm}

\myparagraph{3D Geometric Control} Our flexible framework enables us to create novel scenes by manipulating voxels as displayed in Fig.~\ref{fig:e2etest} and Fig.~\ref{fig:edit}. Basically, we can edit the occupied state and semantics of every voxel in our scenes for generation. We highlight that we can create a hinged-articulated truck and an excavator as shown in Fig.~\ref{fig:ctrl_compare}~(d,e). The generated excavator image exhibits a remarkable alignment with the input occupancy that is delineated by a black outline.

% The manipulation is conducted in point space. We first create a set of point clouds with desirable semantic label, or retrieve point clouds from 3D assets. Then we add the new point cloud to the original scene point cloud and voxelize the point cloud to form the MPI. 

\begin{table*}[b]
\resizebox{0.85\textwidth}{!}{
\begin{floatrow}
\hspace{-40pt}
\capbtabbox{
    % \resizebox{0.3\textwidth}{!}{
\begin{tabular}[width=0.9\textwidth]{lccc}
\toprule
\textbf{Method} & \textbf{Condition Type} & \textbf{FID} \\
\midrule
 BEVGen~\cite{swerdlow2024street}   &   BEV map
  &   25.54    \\ 
 BEVControl~\cite{yang2023bevcontrol}  &   BEV map
   &   24.85   \\ 
 DriveDreamer~\cite{wang2023drivedreamer}  &   Box + FoV map   &   52.60   \\ 
 MagicDrive~\cite{gao2023magicdrive}  &   Box + BEV map
   &   16.20   \\ 
 Panacea~\cite{wen2023panacea} &   Box + FoV map    &   16.96   \\ 
 Ours &   3D Semantic MPI  &   \textbf{14.75}    \\
\bottomrule
\end{tabular}
    % }
}{
 \caption{Comparison of FID with previous methods \\ on the nuScenes dataset.}
 \label{tab:comparefid}
}
\hspace{-8pt}

\capbtabbox{
    \begin{tabular}[width=0.9\textwidth]{cccc|c}
    \toprule
\multicolumn{1}{c}{\textbf{MPI Encoder}} & \multicolumn{3}{|c|}{\textbf{Reweighing Method}} & \multicolumn{1}{c}{\textbf{Metric}} \\
% \cmidrule{1-2} 
% \cmidrule{3-5} 
\midrule

   \multicolumn{1}{c|}{ Design } & Progressive &  Depth  & CBGS  & mIoU \\
     \midrule
         3×3& - &  - &   -&    21.96   \\
          1×1 & - &  - & -  &    23.05   \\
          1×1 & \cmark & -  &  - &    23.63   \\
          1×1 & \cmark &  \cmark & -  &    24.40   \\
         1×1 & \cmark &  \cmark &  \cmark &    25.50   \\
    \bottomrule
    \end{tabular}
}{
 \caption{Ablation of different designs of the MPI encoder and reweighing methods.}
 \label{tab:ablate}
 \small
}
\end{floatrow}
}
\end{table*}

    \vspace{-1mm}

\myparagraph{Long-tailed Scene Generation} The flexibility of 3D semantic MPI has conferred significant advantages upon our approach. In the following, we create long-tail scenes that rarely occur in our real world for evaluation. In Fig.~\ref{fig:e2etest}, we show that we manually add parallel traffic cones in front of the ego vehicle. This scene has never happened in the training dataset, but our geometric controllability provides us the capability to create such data. We then use the created scene to test autonomous driving systems such as end-to-end planner VAD~\cite{jiang2023vad} to validate its effectiveness. In this case, VAD successfully predicts correct waypoints with the high-level command `turn left'. Moreover, in appendix Sec.~\ref{sec:longtaileval}, we generate long-tailed scenes with extreme weather such as snow and sandstorms, and evaluate perception model on it to examine its generalizability of rare weather.

% Detailed settings and results are provided in the appendix. 

% More examples are provided in the appendix. 

\myparagraph{Comparison with Baselines}
In Fig.~\ref{fig:ctrl_compare}~(a), we visualize a comparison with ControlNet. We find that ControlNet struggles to distinguish the overlapping instances in 2D-pixel space. This leads to the two parked cars being merged into a single car with incorrect pose. In contrast, our 3D semantic MPIs contain more than 2D semantic mask, but also account for complete scene geometry with occluded parts. Together with our proposed MPI encoder and reweighing strategy, our framework yields a realistic image generation with high-quality label alignment. More comparison is provided in Sec.~\ref{sec:suppcompare}.

\subsection{Quantitative Results}

\myparagraph{Recognizability, Realism and Controllability Evaluation} To evaluate whether our generated images aligned with given annotations, we provide Gen experiment in Tab.~\ref{tab:base_main}. Using the annotation of val set, we synthesize a copy of val set's images, then use perception model trained on real training set to perform evaluation. The performance will be more effective as it is close to the oracle performance. We find that local method (ControlNet) perform better than global method (MagicDrive). Furthermore, \methodnamesmall generalizes the locality for 3D conditioning to yield better performance.

% Our method achieve a high degree of improvement compared with prior works.

\myparagraph{Data Augmentation for 3D Occupancy Prediction} Notably, we conduct experiments using our synthesized dataset to enhance the real training set in Tab.~\ref{tab:base_main}. We first use the occupancy labels from training set to create a synthetic training set. Then we modify the loading pipeline in perception model to randomly sample images from real dataset or synthetic dataset and train network from scratch. Therefore, our approach preserves the inherent training dynamics of the neural network by solely modifying the training images, without any alteration to the number of training iterations or epochs. As MagicDrive-Aug exhibits numerical overflow when training FB-Occ, which may attributed to unsatisfactory recognizability, we have to omit it and only provide MagicDrive-Gen experiments.

% As MagicDrive-Aug exhibits numerical overflow when training FB-Occ, we only provide MagicDrive-Gen experiments. 

As shown in Tab.~\ref{tab:base_main}, where \methodnamesmall-Aug denotes the augmentation experiments using our generated dataset, shows a satisfactory improvement over the prior state of the art. We emphasize that surpassing the performance of the original dataset is not the primary objective of our work; rather, it is an ancillary benefit that emerges from our framework for geometry-controlled generation. 

% Future work can extend our work to generate more images with rare classes to validate the improvement of long-tailed classes.

% As for MagicDrive, it can be difficult to condition the large-scale occupancy input into 1D embedding. 

% The ControlNet+depth can aid with this problem since the two cars are with different depth. However, we find that the the edges of objects from ControlNet+depth tend to become blurred. 

% Compared with ControlNet, we provide condition information with 3D information, thus enabling generation with 3D reasoning. Compared with ControlNet with depth, our proposed occupancy encoder is more effective in providing the conditional signal. Besides, 3D semantic MPIs contain more than 2.5D information in semantic map+depth, but with occluded parts. 

\myparagraph{Ablations}\label{sec:ablate} In Tab.~\ref{tab:ablate}, we present ablation studies across several design spaces of our model, analogous to the Gen experiment in Tab.~\ref{tab:base_main}. We find that our designed MPI encoder of 1×1 conv have significant improvement when compared to the conventional 3×3 conv approach. Besides, our proposed three types of reweighing methods demonstrate a consistent improvement over the baseline. As a result, the improved image quality and label alignment enable higher precision in downstream tasks.

    \vspace{-1mm}

\section{Limitation and Broader Impacts}\label{sec:limitation}

    \vspace{-1mm}

\myparagraph{Layout Genereation} Our method is restricted in a conditional generation framework that should have a conditional input at first. Our condition signal is from the original dataset annotation. Thus most of the augmented data is generated using the same occupancy layout, or with minimal human editing. Future research can incorporate the recent research~\cite{lee2024semcity,urbandiff,lu2023wovogen,cottraffic,wu2024blockfusion} that generates occupancy and traffic descriptions of the scenes to synthesize images with novel occupancy or traffic layouts.

    \vspace{-1mm}

\myparagraph{Closed-loop Simulation} Given the underlying diverse and controllable image generation of our method, it would be advantageous and valuable to extend our work to a broader domain such as closed-loop simulation~\cite{ljungbergh2024neuroncap,yang2023unisim}, to enable high-fidelity autonomous systems testing. This line of work can be conducted by utilizing motion conditions to generate future frames as in world model~\cite{yang2023learning,wang2023driving,lu2023wovogen}, or by explicitly modeling scene graph as in the case of UniSim\cite{yang2023unisim,Ost_2021_neuralscene} and NeuroNCAP~\cite{ljungbergh2024neuroncap}.

% \myparagraph{Instance-level Prompt Control} In our setting, we perform class-condition image generation.

    \vspace{-1mm}

\myparagraph{Long-tailed Scene Generation} In this paper, we only investigate a limited number of long-tailed scene generation and corner case evaluations such as rare layout in Fig.~\ref{fig:e2etest} and extreme weather in Sec.~\ref{sec:longtaileval}. Future work can extend our framework to \textbf{(i)} Synthesize more samples for tail classes to boost performance. \textbf{(ii)} Generate or replicate large-scale databases of corner cases~\cite{li2022coda} for robust perception.

\section{Conclusion}\label{sec:conclusion}

    \vspace{-1mm}

In this paper, we propose \methodname, an innovative image generation framework that is empowered with geometry-controlled capabilities using occupancy. We introduce a novel 3D representation, 3D semantic MPIs, to address the critical challenge of how to efficiently encode occupancy. This representation not only preserves the authentic and complete 3D geometry details with semantics, but also provides a spatial-align feature representation for 2D diffusion models. With this property, our method enjoys photorealistic appearances and fine-grained 3D controllability, serves as a generative data engine to enable a broad range of applications. Extensive experiments demonstrate that our synthetic data facilitate the training for perception models on occupancy prediction, and provide valuable corner case evaluation in a simulated world. 

% We address the critical challenge of how to efficiently encode occupancy by introducing a novel 3D representation, 3D semantic MPIs.

{\small
\bibliographystyle{plain}
\bibliography{parts/07_reference}
}

\newpage
\appendix
\section*{\LARGE \centering Appendix}
% \vspace{12mm}

In the appendix, we provide the following content:

% Sec.~\ref{sec:stategeo}: Statement of Geometric Control and our Extension. Sec.~\ref{sec:longtaileval}: Long-Tailed Scene Evaluation. Sec.~\ref{sec:ablateplane}: Ablation of plane number in MPIs. Sec.~\ref{sec:suppcompare}: Additional Qualitative Comparison. Sec.~\ref{sec:suppvideo}: Results of Video Generation. Sec.~\ref{sec:newcam}: Generalize to New Cameras. Sec.~\ref{sec:influence_amount}: The Impact of the Amount of Augmented Data. Sec.~\ref{sec:failcase}: Failure Cases. 

\begin{table}[h]
\begin{tabular}{|l|l|}
\hline
Sec.~\ref{sec:stategeo}: Statement of Geometric Control. & Sec.~\ref{sec:suppvideo}: Results of Video Generation.                        \\ \hline
Sec.~\ref{sec:longtaileval}: Long-Tailed Scene Evaluation.                 & Sec.~\ref{sec:newcam}: Generalize to New Cameras.                             \\ \hline
Sec.~\ref{sec:ablateplane}: Ablation of plane number in MPIs.              & Sec.~\ref{sec:influence_amount}: Impact of Amount of Augment Data. \\ \hline
Sec.~\ref{sec:suppcompare}: Additional Qualitative Comparison.             & Sec.~\ref{sec:failcase}: Visualization of Failure Cases.                                       \\ \hline
\end{tabular}
\end{table}

\maketitle
% In this \textbf{supplementary document}, we first provide

\section{Statement of Geometric Control}\label{sec:stategeo}

In our paper, we refer the geometric controllable generation as using a voxel grid in 3D space to control the image generation. Although the voxel is a quantized representation of the 3D world, when the resolution goes larger, it can already faithfully represent the geometry detail of scenes. Currently, we are limited by the precision of ground truth labels. The $0.2m$ occupancy grid is a tensor of 500×500×40 that cover a space in x-axis spanning $[-50m, 50m]$, y-axis spanning $[-50m, 50m]$, z-axis spanning $[-5m, 3m]$. In the future, we plan to explore a higher resolution of geometric control to refine our generation.

Except for occupancy, several other 3D representations can be expressed by 3D semantic MPI, such as mesh, dense point clouds, and even 3D boxes or HD maps. The underlying mechanism is to cast several slices of multi-plane images at different depths to retrieve geometric information. Thus, our 3D semantic MPI can be regarded as a general 3D conditioning representation to benefit a wide spectrum of practical systems. These encompass but are not limited to 3D generation such as text2room~\cite{hollein2023text2room}, RoomDreamer~\cite{song2023roomdreamer}, WonderJourney~\cite{yu2023wonderjourney}, and LucidDreamer~\cite{chung2023luciddreamer}, each of which stands to benefit from the rich geometric context provided by our approach.

% ControlRoom3D~\cite{schult2023controlroom3d},

\section{Long-Tailed Scene Evaluation}\label{sec:longtaileval}

In this section, we explore to use \methodnamesmall to create long-tailed scenes for downstream evaluation. This also stands for evaluating our model using several corner cases. Similar to the SytheOcc-Gen experiment in Tab.~\ref{tab:base_main}, we generate a synthetic validation set but use prompts control to manipulate weather patterns or the intensity of illumination.

\begin{figure*}[!h]
    \centering
    \includegraphics[width = \textwidth, trim = 0 0 0 0, clip]{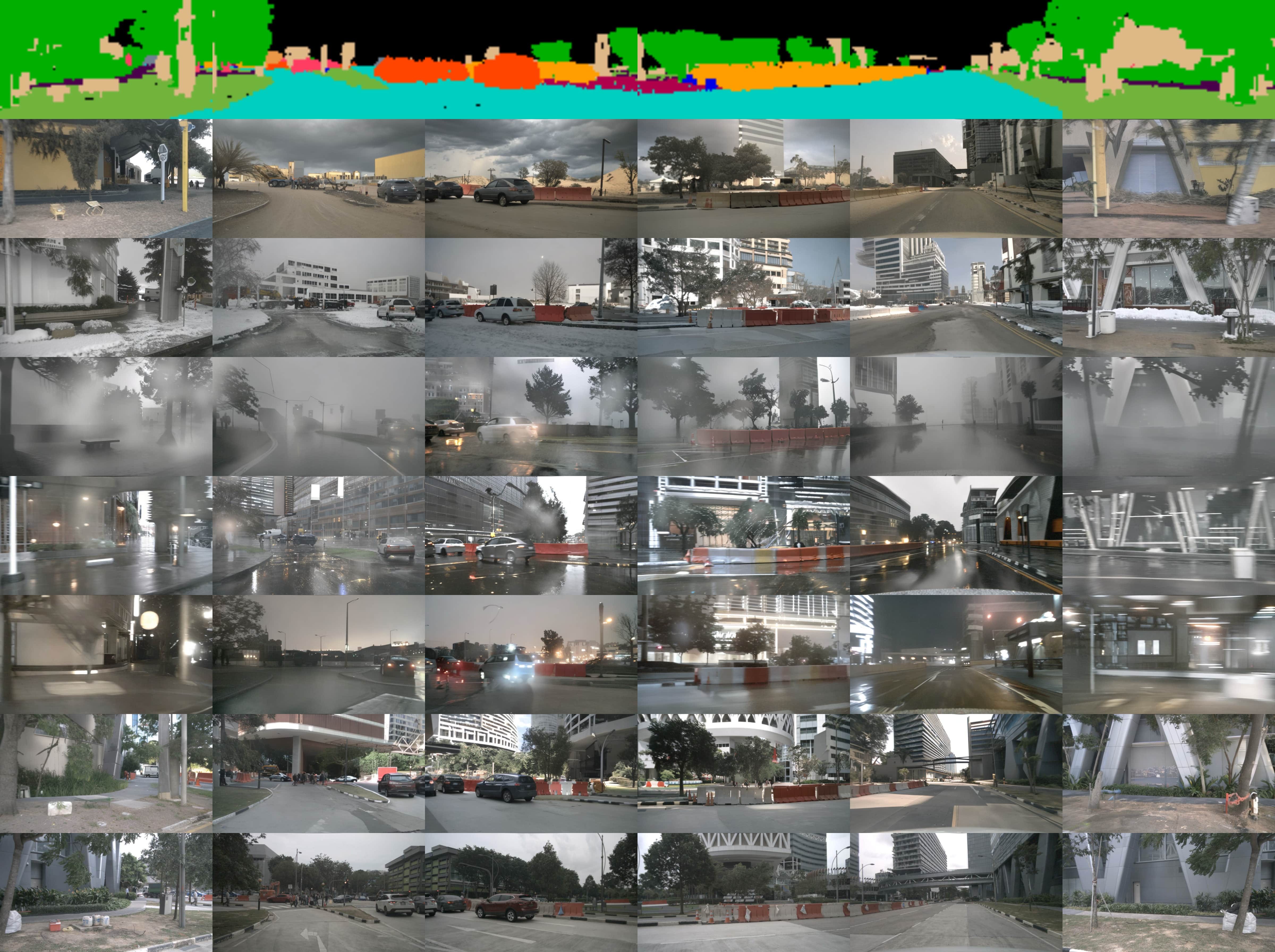}
    \caption{From top to bottom, we display images of fusion of 3D semantic MPI, synthesized images of sandstorm, snow, foggy, rainy, day night, day time, and ground truth.}
    % \vspace{-1mm}
    \label{fig:suppweather}
\end{figure*}

As depicted in Fig.~\ref{fig:suppweather}. We create a variety of weather conditions including sandstorms, snow, foggy, rainy, day night, and day time. The motivation behind the creation of these scenes lies in their extreme rarity compared to the ordinary scenes we have captured. The generation of such data is of significant value, as it aids in addressing the long-tailed distribution of scenes, thereby enriching the diversity of our dataset. More visualization is provided in Fig.~\ref{fig:suppweather1} to Fig.~\ref{fig:suppweather3}.

In Tab.~\ref{tab:snoweval}, we observe that all kinds of extreme weather lead to a degradation in performance. This observation underscores the limitations of the perception model in terms of its generalizability to infrequent weather scenarios. Among them, we find that foggy, rainy, and day night exert the most severe impact, as they contribute to a large reduction in visibility as shown in  Fig.~\ref{fig:suppweather}. To improve the generalizability to handle various weather conditions, future work can leverage our generated data to cover the long-tailed scenes, or use adversarial search to find severe scenes based on our framework.

\begin{table*}[h]
    \centering
    % \vspace{-1.0mm}
        \resizebox{0.8\textwidth}{!}{
        \begin{tabular}[width=\linewidth]{lcccccc}
        \toprule
        \textbf{Scenes} & \textbf{Sandstorm} & \textbf{Snow} &  \textbf{Foggy} & \textbf{Rainy} & \textbf{Day night} & \textbf{Day time} (raw data) \\
        \midrule
        % \midrule
        FB-Occ mIOU	& 22.88 & 18.25 &		10.29 &	9.71 & 9.95 &	25.50 \\
        \bottomrule
        \end{tabular}
        % \vspace{-.7em}
        }
    \caption{Experiments of downstream evaluation on long-tailed scenes with extreme weather.}
    % \vspace{-.1em}
    \label{tab:snoweval}
\end{table*}

\begin{figure*}[!t]
    \centering
    \includegraphics[width = \textwidth, trim = 0 0 0 0, clip]{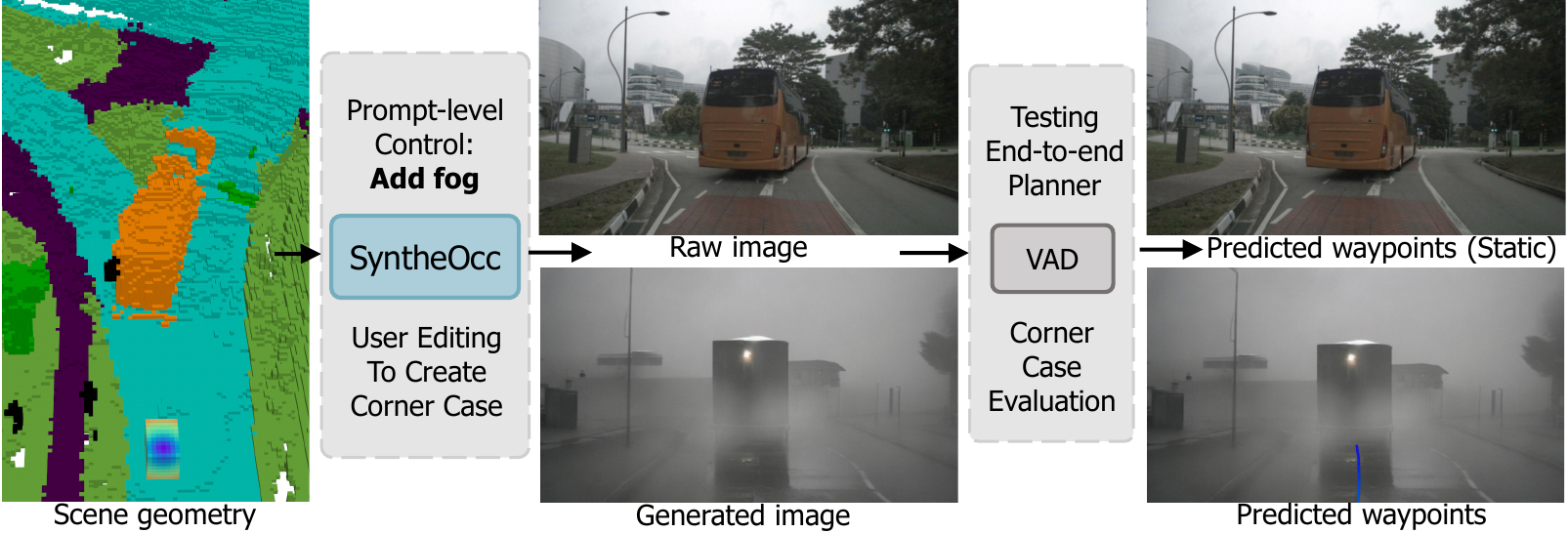}
    \caption{Use \methodname to create long-tailed scenes for testing. \textbf{Top}: In the ordinary scene of a bus placed in front of the ego vehicle, the end-to-end planner VAD~\cite{jiang2023vad} predicts future waypoints without movement, thus not plotted in the image. \textbf{Bottom}: By harnessing the prompt-level control in our framework, we simulate a scene with the same layout but filled with fog. VAD predicts wrong waypoints that will collide with the bus.}
    \vspace{-1mm}
    \label{fig:suppe2etest}
\end{figure*}

Furthermore, we perform long-tailed scene evaluation in Fig.~\ref{fig:suppe2etest}. We display the failure of the downstream model VAD~\cite{jiang2023vad} in our synthetic long-tailed scene. In this case, we simulate a foggy environment that the dense fog obscures the majority of the ego view. Our experiment reveals that due to the lack of training images of foggy scenes, VAD erroneously predicts waypoints that would result in a collision with the bus. This experiment elucidates the boundaries and failure cases of the VAD model~\cite{jiang2023vad}. It exposes the limitations of the system under certain conditions, thereby providing insights into scenarios where the model's performance may be compromised.

\vspace{-2mm}

\begin{figure*}[!h]
    \centering
    \includegraphics[width = \textwidth, trim = 0 0 0 0, clip]{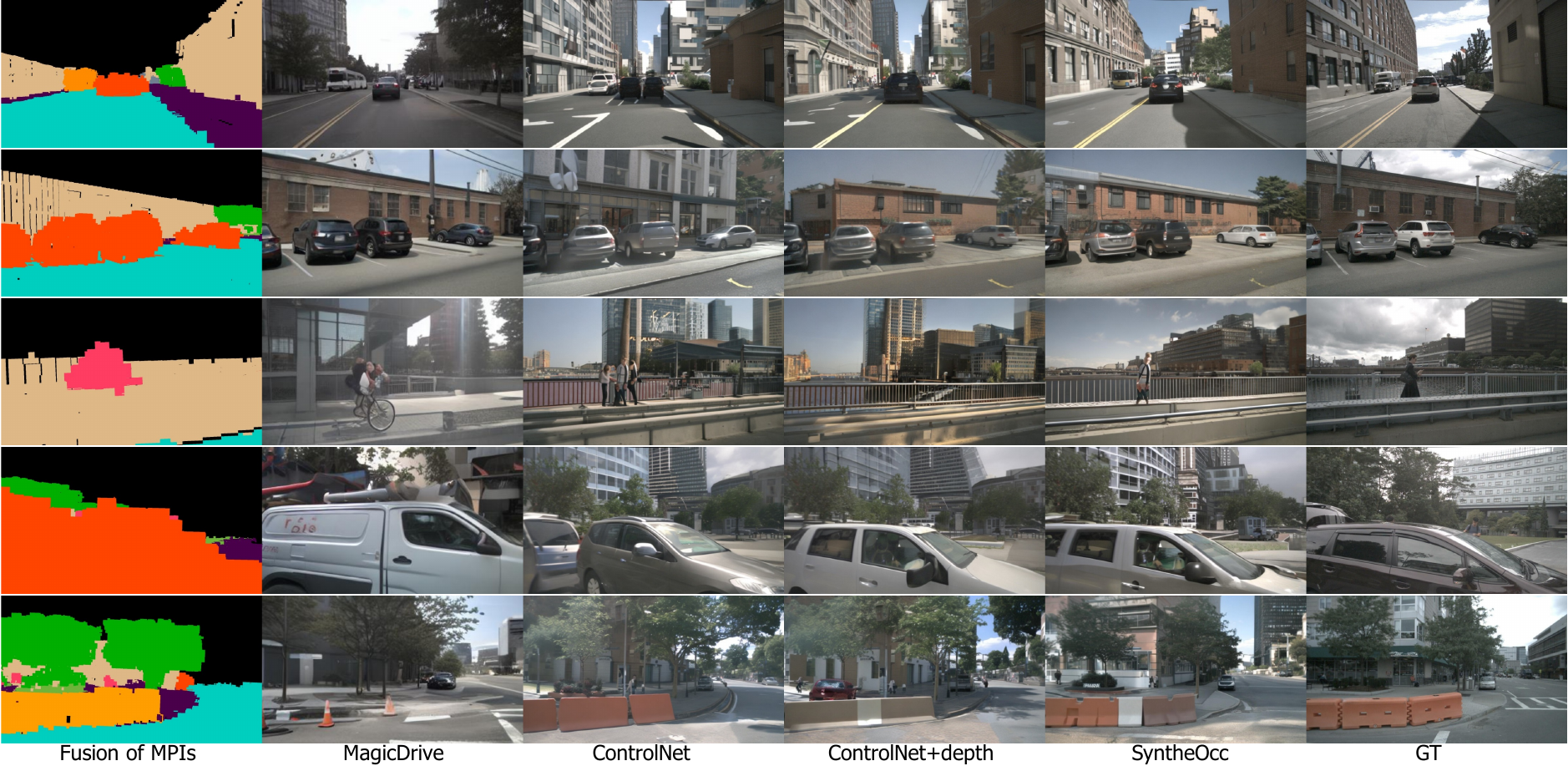}
    \caption{Comparison with baselines.}
    % \vspace{-1mm}
    \label{fig:suppinfer}
\end{figure*}

\section{Ablation of plane number of MPIs}\label{sec:ablateplane}

In our proposed 3D semantic MPIs, the number of planes is a hyperparameter that affects the precision of 3D representation. The plane number can be regarded as the 3D resolution in depth axis. The larger the plane number, the MPI will contain more details. We find that an increase in the number of planes is associated with improved accuracy in downstream tasks. This finding denotes that more condition information leads to better downstream task performance.

\begin{table*}[h]
    \centering
    % \vspace{-1.0mm}
        \resizebox{0.4\textwidth}{!}{
        \begin{tabular}[width=\linewidth]{lccc}
        \toprule
        \textbf{Number of Planes} & \textbf{96} & \textbf{128} & \textbf{256} \\
        \midrule
        % \midrule
        FB-Occ mIOU	& 23.36 &	24.28 &	25.50 \\
        \bottomrule
        \end{tabular}
        % \vspace{-.7em}
        }
    \caption{Ablation of the number of multi-plane images.}
    % \vspace{-.1em}
    \label{tab:ablateplane}
\end{table*}

% \begin{table*}[b]

% \resizebox{0.85\textwidth}{!}{
% \begin{floatrow}
% \hspace{-40pt}

% \capbtabbox{
%     % \resizebox{0.3\textwidth}{!}{
% \begin{tabular}[width=0.9\textwidth]{lcc}
% \toprule
% \textbf{Number}  & \textbf{mIOU} \\
% \midrule
%  96 &   25.54    \\ 
%  128 &   24.85   \\ 
%  256   &   52.60   \\ 
% \bottomrule
% \end{tabular}
%     % }
% }{
%  \caption{Ablation of the number of multi-plane images.}
%  \label{tab:ablateplane}
% }
% \hspace{-8pt}

% \capbtabbox{
%     \begin{tabular}[width=0.9\textwidth]{lcc}
% \toprule
% \textbf{Scenes}  & \textbf{mIOU} \\
% \midrule
%  Snow    &   25.54    \\ 
%  Sandstorm   &   24.85   \\ 
%  Rainy     &   52.60   \\ 
%  Foggy & 16.20   \\ 
%  Day night  & 16.96   \\ 
%  Day time   &   \textbf{14.75}    \\
% \bottomrule
% \end{tabular}
% }{
%  \caption{Experiments of long-tailed scene generation with extreme weather.}
%  \label{tab:snoweval}
%  \small
% }
% \end{floatrow}
% }
% \end{table*}

% \section{Ablation}
\section{Qualitative Comparison with Baselines and SOTA}\label{sec:suppcompare}

In Fig.~\ref{fig:suppinfer}, we conduct a qualitative comparison of our method against MagicDrive, ControlNet, and ControlNet+depth. We find that all the methods display a satisfactory image quality, as they build upon the foundation of the stable diffusion model. The generation of MagicDrive fails to synthesize barriers as shown in the bottom row. ControlNet struggles to generate objects with the correct pose solely from only 2D conditions as shown in the second row. ControlNet+depth, a degradation of our method, an enhancement over ControlNet in terms of alignment, nevertheless suffers from a loss of finer detail in scenes with heavy occlusion, as shown in the human of the third row. Our method, in contrast, aims to address these challenges and provide a more nuanced and accurate generation of complex scenes.

    \vspace{-2mm}

\section{Extend to Video Generation}\label{sec:suppvideo}

    \vspace{-2mm}

As described in the main paper Sec.~\ref{sec:attention}, we further extend the cross-view attention to cross-frame attention to perform video generation. Our generation results are Fig.~\ref{fig:video_vis}, Fig.~\ref{fig:video_vis1} and Fig.~\ref{fig:video_vis2}. Our implementation is adopted from MagicDrive~\cite{gao2023magicdrive} which is similar to Tune-a-video~\cite{wu2023tune}. The formulation of cross-frame attention is:
\begin{align}
    \texttt{Attention}(Q, K, V) = \texttt{softmax}(\begin{matrix}\frac{QK^{T}}{\sqrt{d}}\end{matrix}) \cdot V\text{,}
    \\
    % {h}_{cross\_view}^{v} = {h}_{in}^{v}  + \sum_{i\in\{l,r\}}\texttt{Attention}(Q_{in}, K_{i}, V_{i})\text{,} \\
    {h}_{out} = {h}_{in}  + {\textstyle\sum_{i\in\{f,h\}}}\texttt{Attention}(Q_{in}, K_{i}, V_{i})\text{,}
    \label{equ:cf-attn}
\end{align}
where $f$, and $h$ are the camera view of future and history frames. $Q_{in}$ and ${h}_{in}$ denotes the query and the hidden state of input view. We train our model in a two-stage pipeline. We first train the MPI encoder and cross-view attention in a multi-view image generation setting. Then we train cross-frame attention and freeze other components in a video generation setting.

     % {h}_{out} = {h}_{in}  +  {\textstyle \sum_{i\in\{l,r\}}^{}} \texttt{Attention}(Q_{in}, K_{i}, V_{i})\text{,}

In practice, we use the keyframe annotation of the nuScenes dataset to train our video model. We start with our pretrained MPI encoder and cross-view attention and only train our cross-frame attention while keeping others frozen. We employ a sequence of 7 frames as a batch, resulting in a batch size of 42 images for the training process. 

Given that our primary contribution does not lie in video generation, this experiment serves as a proof of concept, demonstrating the potential of our framework. Future research may extend our methodology to facilitate the generation of longer video sequences, thereby expanding the scope and applicability of our framework.

    \vspace{-2mm}

\section{Generalize to New Cameras}\label{sec:newcam}

    \vspace{-2mm}

\begin{figure*}[!t]
    \centering
    \includegraphics[width = \textwidth, trim = 0 0 0 0, clip]{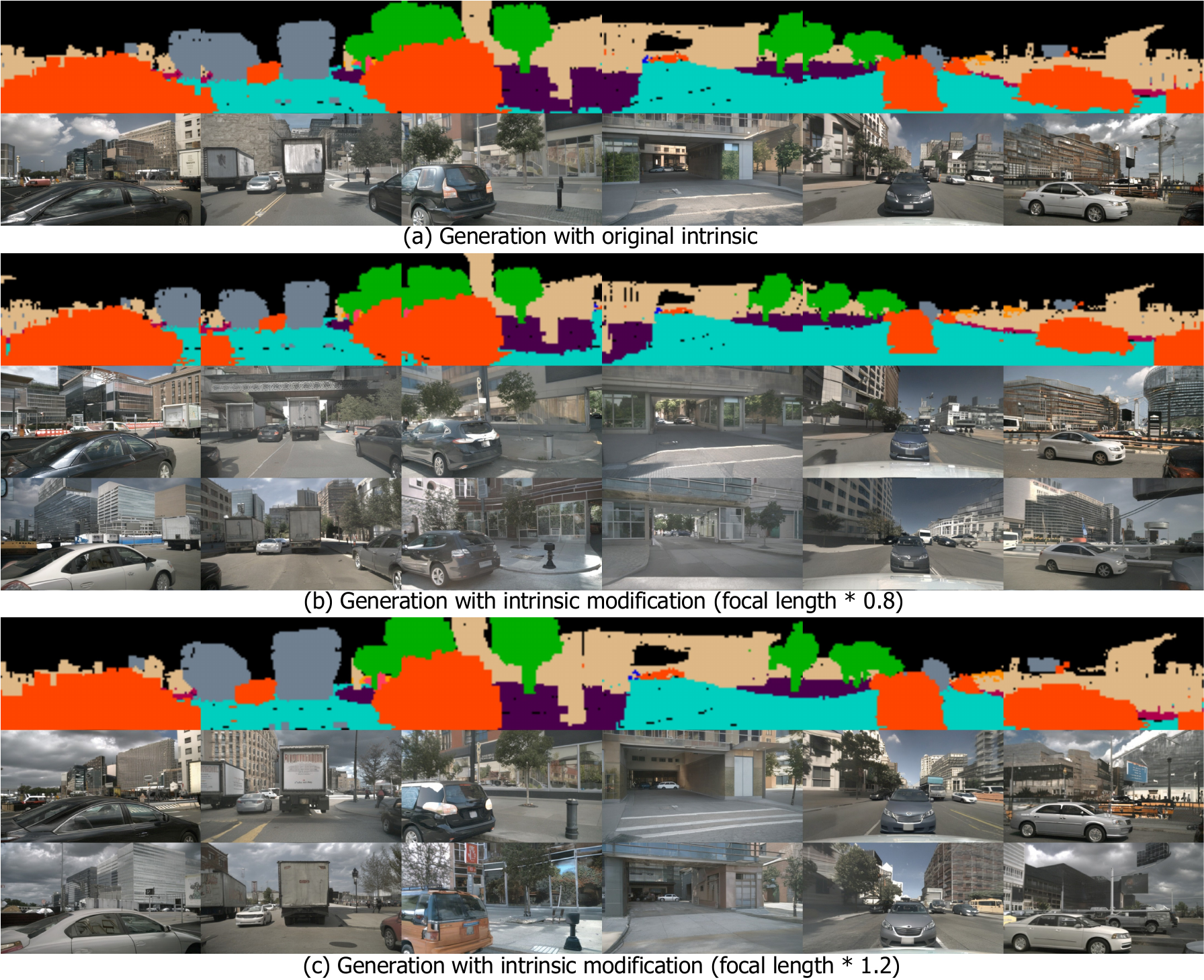}
    \caption{We demonstrate the generalizability of \methodnamesmall to new camera intrinsic. We multiply factors to the focal length while keeping the resolution the same. In (b,c), focal length $\times 0.8$ denotes a camera with a larger field of view similar to zoom out, focal length $\times 1.2$ denotes a camera with a smaller field of view similar to zoom in. }
    % \vspace{-1mm}
    \label{fig:newcam}
\end{figure*}

\begin{figure*}[!h]
    \centering
    \includegraphics[width = \textwidth, trim = 0 0 0 0, clip]{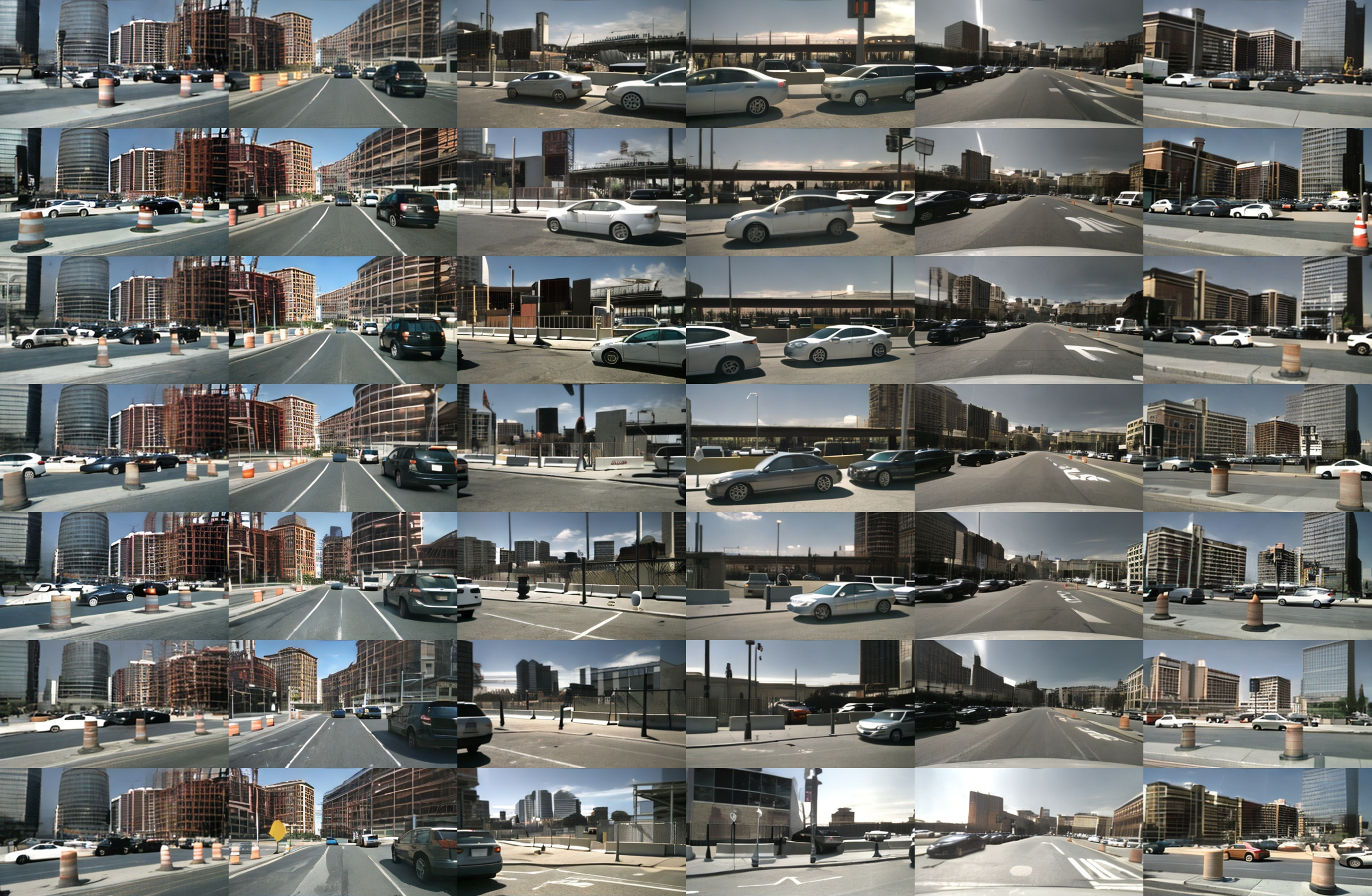}
    \caption{Video generation results. In the temporal progression, the distant buildings maintain a high degree of consistency, and objects retain their identical shapes and textures across different views and frames.}
    % \vspace{-1mm}
    \label{fig:video_vis}
\end{figure*}

\begin{figure*}[!h]
    \centering
    \includegraphics[width = \textwidth, trim = 0 0 0 0, clip]{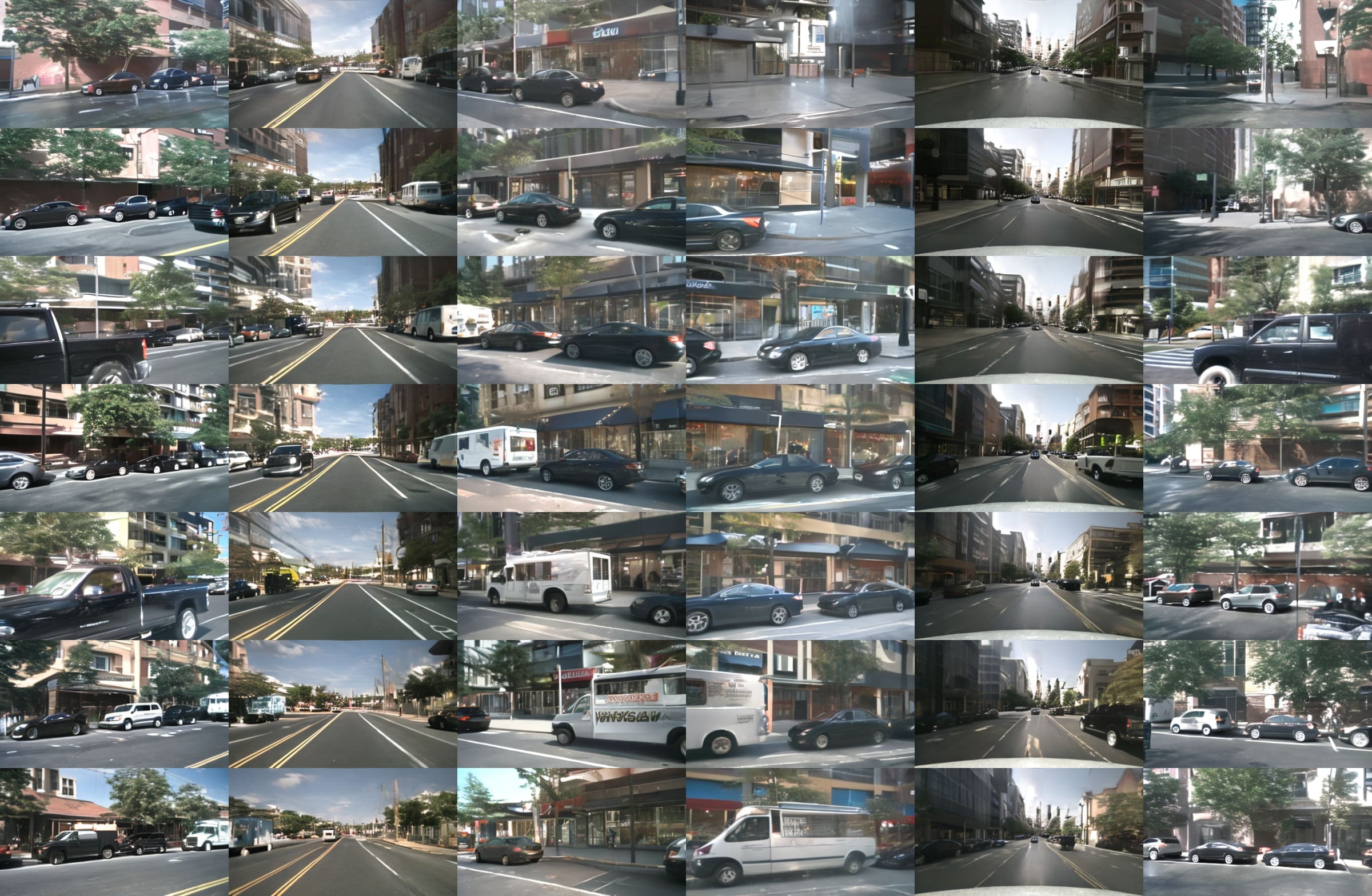}
    \caption{Video generation results of large dynamics scenes. The white car comes across different views and frames depicting consistent shapes with only a slight appearance change. }
    % \vspace{-1mm}
    \label{fig:video_vis1}
\end{figure*}

\begin{figure*}[!h]
    \centering
    \includegraphics[width = 0.98\textwidth, trim = 0 0 0 0, clip]{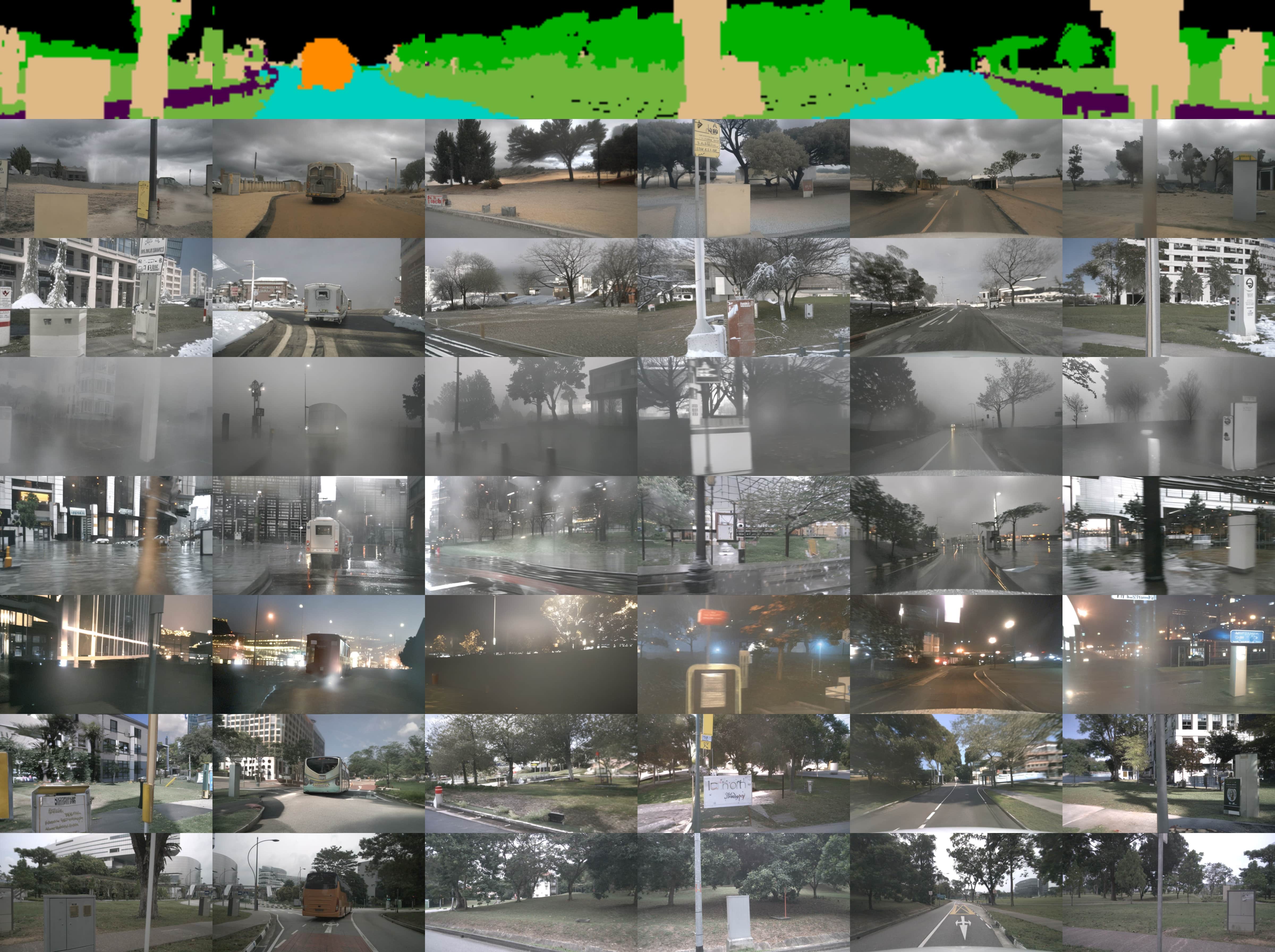}
    \caption{From top to bottom, we display images of fusion of 3D semantic MPI, synthesized images of sandstorm, snow, foggy, rainy, day night, day time, and ground truth.}
    % \vspace{-1mm}
    \label{fig:suppweather1}
\end{figure*}

\begin{figure*}[!h]
    \centering
    \includegraphics[width = 0.98\textwidth, trim = 0 0 0 0, clip]{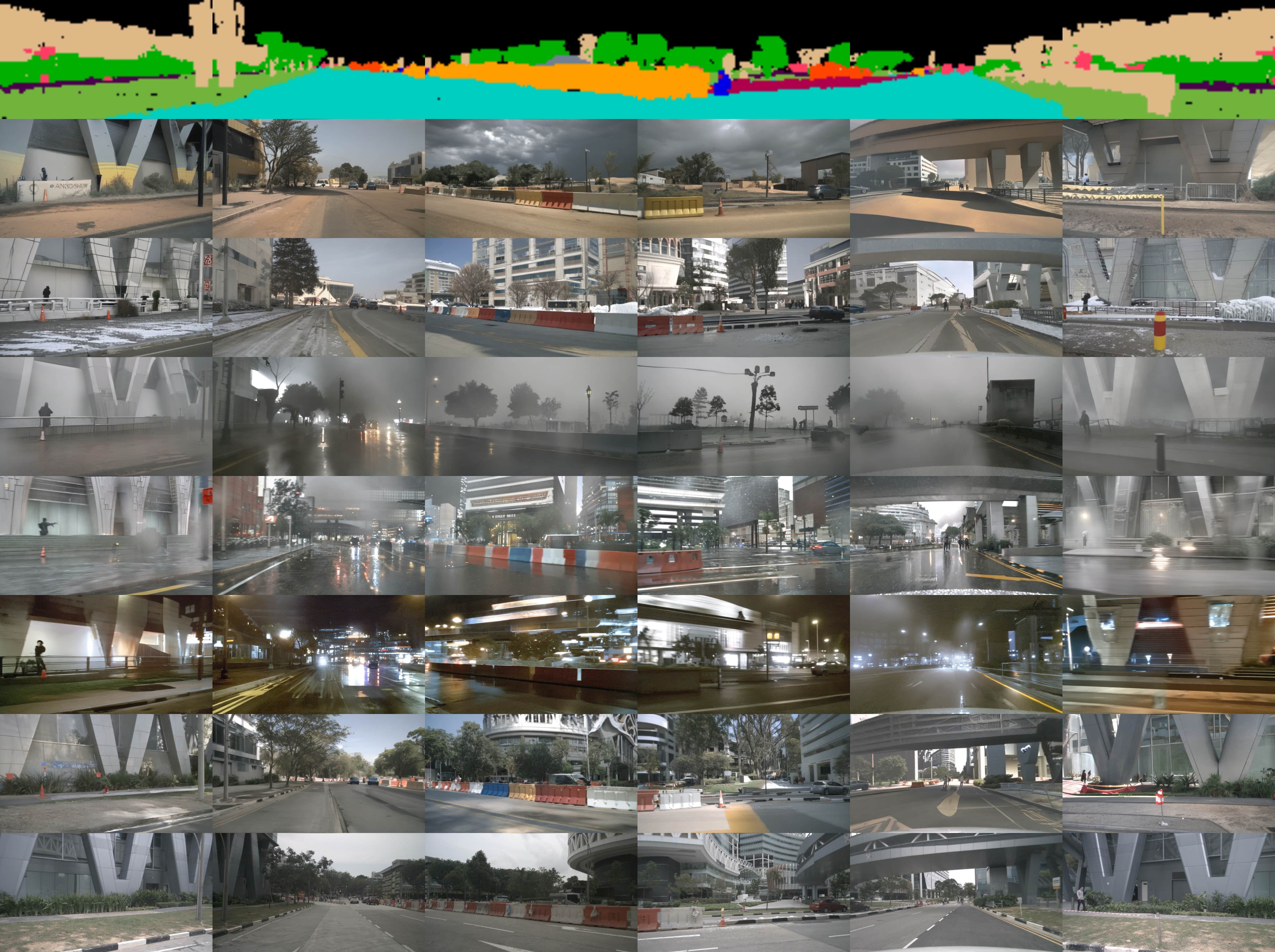}
    \caption{Weather variation. Same structure with Fig.~\ref{fig:suppweather1}.}
    % \vspace{-1mm}
    \label{fig:suppweather3}
\end{figure*}

In this section, we investigate the adaptability of our method to a new set of cameras with different intrinsic. Given that our training set has a fixed camera intrinsic and extrinsic, generalizing to novel cameras indicates that our approach possesses robust generalization capabilities. As shown in Fig.~\ref{fig:newcam}, benefiting from our local type of condition, \methodnamesmall generates images that faithfully align with the new intrinsic, proving that \methodnamesmall do not over-fit certain parameters. Regarding extrinsic parameters, we can cast our MPI at the desirable locations to retrieve geometric information, thus inherently ensuring generalizability without doubt.

    \vspace{-2mm}

\section{The Influence of the Amount of Augmented Data}\label{sec:influence_amount}

    \vspace{-2mm}

As \methodnamesmall is capable of generating an infinite number of synthetic data, we investigate the influence of the amount of augmented data on downstream tasks in Tab.~\ref{tab:ablateamount}. We find that when our augmented data is expanded from one-fold to two-fold of the training dataset, the performance of perception model slightly decreases. This may indicate the generated data has an optimal ratio for downstream tasks. Due to limited computational resources, we only experiment with a limited amount of ratio. Future work can conduct more thorough experiments to find a universal theorem.

\begin{table*}[h]
    \centering
    % \vspace{-1.0mm}
        \resizebox{0.55\textwidth}{!}{
        \begin{tabular}[width=\linewidth]{lccc}
        \toprule
        \textbf{Amount of Augmented Data} & \textbf{0}~(no augmentation) & \textbf{1} & \textbf{2} \\
        \midrule
        % \midrule
        FB-Occ mIOU	& 39.3 &	40.3  &	40.1 \\
        \bottomrule
        \end{tabular}
        % \vspace{-.7em}
        }
    \caption{Ablation of the amount of augmented data.}
    % \vspace{-.1em}
    \label{tab:ablateamount}
\end{table*}

% Due to limited video training images, our method is difficult to generate crowd scenes, where cross-view attention is challenge to 

% We observe a slight identity change of the white car.

\begin{figure*}[!h]
    \centering
    \includegraphics[width = 0.98\textwidth, trim = 0 0 0 0, clip]{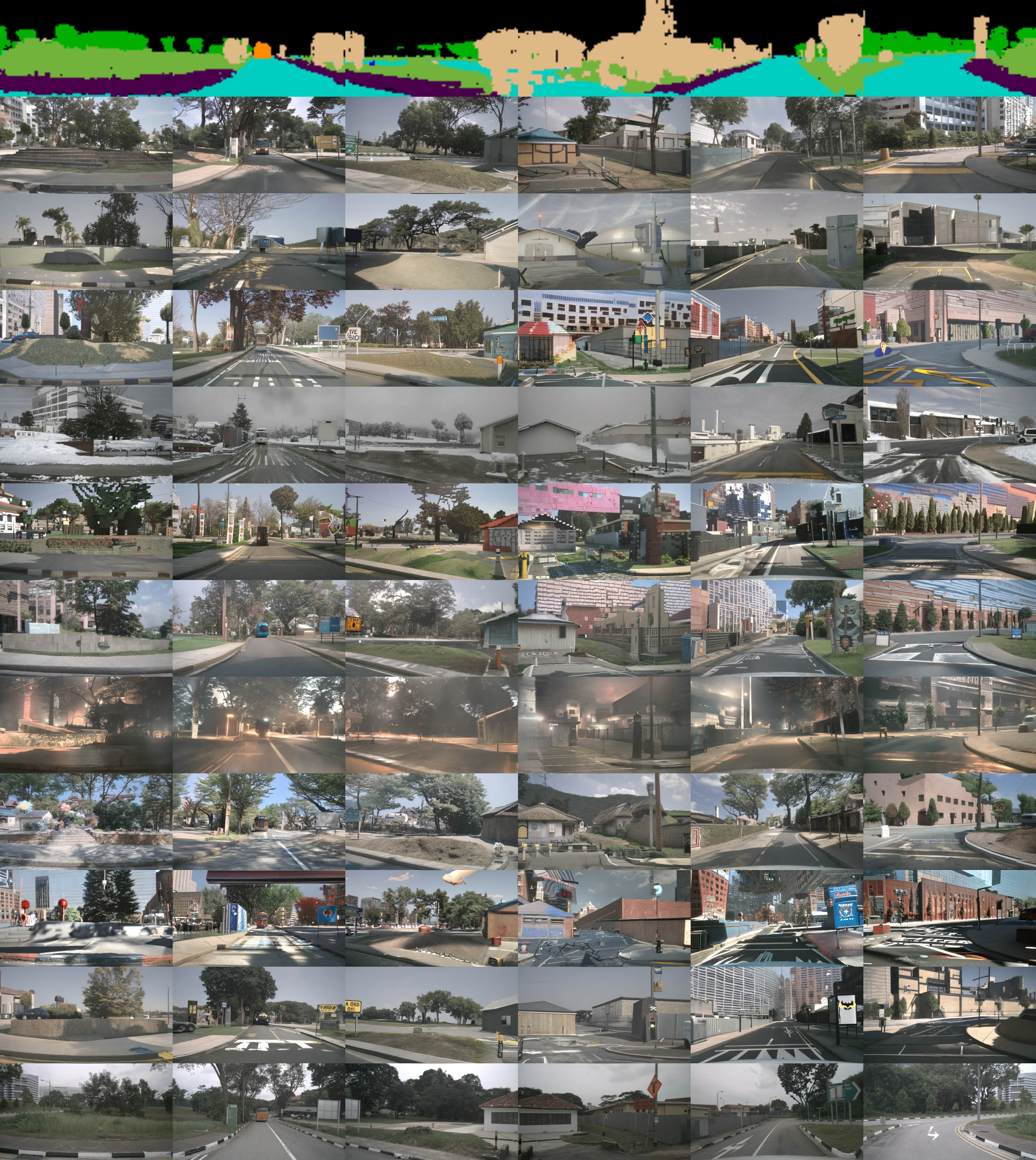}
    \caption{Out of distribution generation. We use prompts to control the high-level appearance of images with specific styles. From top to bottom, we display (1) fusion of 3D semantic MPI. (2) Sunny day. (3) Science fiction style. (4) 8-bit pixel art style. (5) Snowfall. (6) Minecraft style. (7) Pokémon style. (8) Diablo style. (9) Ghibli style. (10) Metropolis style. (11) Gotham style. (12) Ground truth.}
    % \vspace{-1mm}
    \label{fig:suppweather2}
\end{figure*}

\clearpage

\section{Failure Cases}\label{sec:failcase}
\begin{figure*}[!h]
    \centering
    \includegraphics[width = \textwidth, trim = 0 0 0 0, clip]{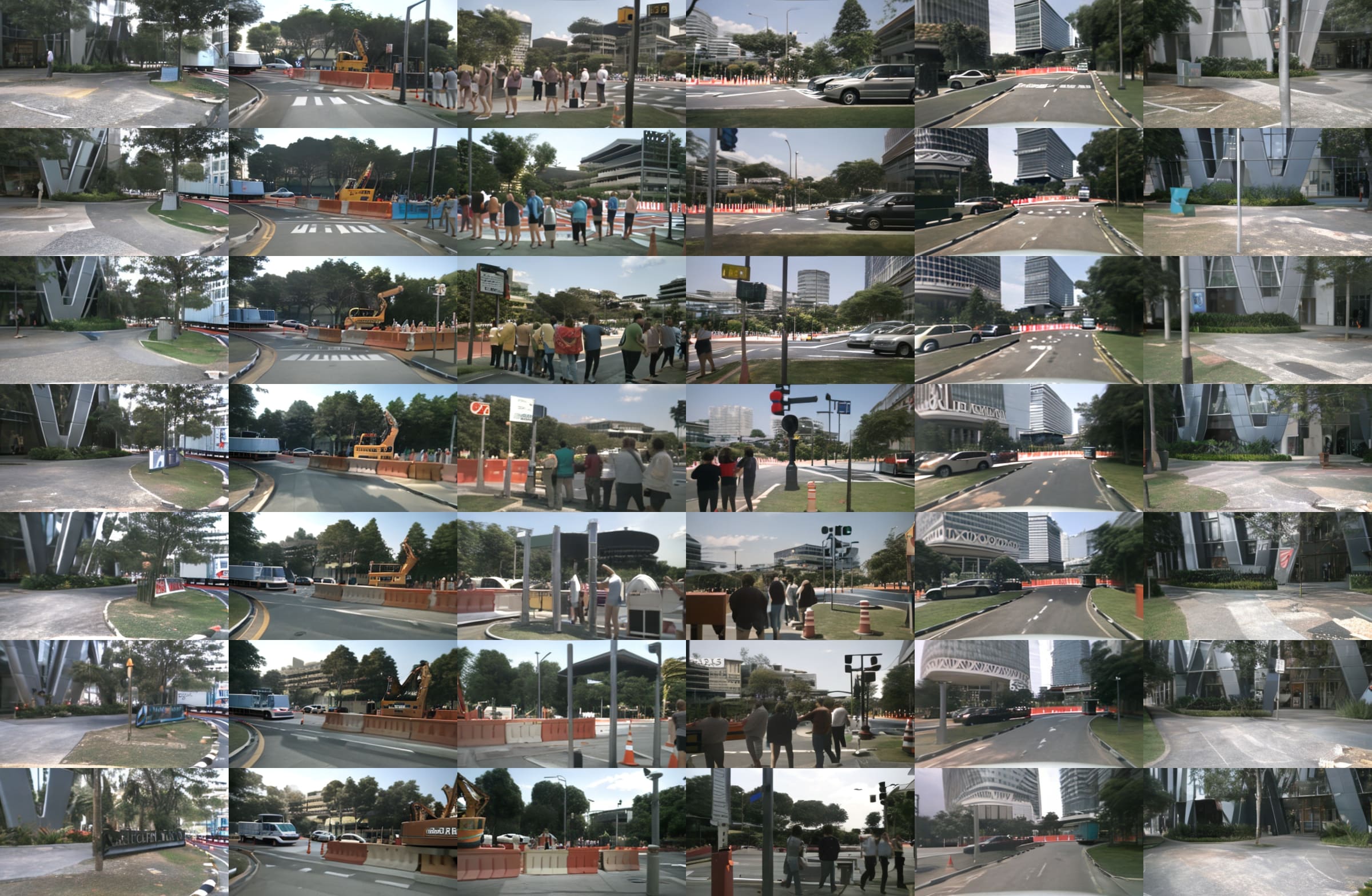}
    \caption{Failure case of video generation results. Our cross-frame attention module is challenging to distinguish a crowd of people across different views and frames.}
    % \vspace{-1mm}
    \label{fig:video_vis2}
\end{figure*}

We display several failure cases of our method. In Fig.~\ref{fig:video_vis2}, we show a crowd scenes. In this scenario, the excessive number of pedestrians presents a challenge to the cross-view attention and cross-frame attention modules. We find our method incapable of discerning individual entities with clarity. Future research can improve the model capacity or enrich high-quality data to mitigate this problem.

\end{document}